%% file: main.tex
\documentclass[lettersize,journal]{IEEEtran}
\usepackage{amsmath,amsfonts}
\usepackage{algorithmic}
\usepackage{algorithm}
\usepackage{array}
\usepackage[caption=false,font=normalsize,labelfont=sf,textfont=sf]{subfig}
\usepackage{textcomp}
\usepackage{stfloats}
\usepackage{float}
\usepackage{url}
\usepackage{verbatim}
\usepackage{graphicx}
\usepackage{cite}
\usepackage{booktabs}
\usepackage{wrapfig,lipsum,booktabs}
\usepackage{multirow}
\usepackage{threeparttable}
\usepackage{amsmath}
\usepackage{xspace}
\usepackage{caption}

\newcommand{\OurMethod}{\emph{BrushEdit}}
\newcommand{\Benchmark}{\emph{BrushBench}}
\newcommand{\TrainingData}{\emph{BrushData}}

\usepackage{colortbl}
\usepackage[dvipsnames]{xcolor}
\definecolor{Red}{RGB}{192, 0, 0}
\definecolor{Blue}{RGB}{12, 114, 186}
\definecolor{Yellow}{RGB}{218, 169, 20}

\definecolor{HighlightBlue}{RGB}{0, 100, 148}
\definecolor{HighlightRed}{RGB}{230, 57, 70}

\definecolor{LightRed}{HTML}{ffe0e0}
\definecolor{LightBlue}{HTML}{def5ff}
\definecolor{LightYellow}{HTML}{FFF6DB}
\definecolor{LightGreen}{HTML}{eff9f0}

\begin{document}

\title{BrushEdit: All-In-One\\Image Inpainting and Editing}


\author{
\textbf{Yaowei Li}$^{1*}$ 
~~
\textbf{Yuxuan Bian}$^{3*}$ 
~~
\textbf{Xuan Ju}$^{3*}$  
~~
\textbf{Zhaoyang Zhang}$^{2\ddagger}$ \\
~~
\textbf{Junhao Zhuang}$^{4}$
~~
\textbf{Ying Shan}$^{2\clubsuit}$ 
~~
\textbf{Yuexian Zou}$^{1\clubsuit}$ 
~~
\textbf{Qiang Xu}$^{3\clubsuit}$ 
~~
\\
$^{1}$Peking University
$^{2}$ARC Lab, Tencent PCG
$^{3}$The Chinese University of Hong Kong
$^{4}$Tsinghua University \\ \quad
$^{*}$Equal Contribution 
~~
$^{\ddagger}$Project Lead
~~
$^{\clubsuit}$Corresponding Author\\
Project Page: \url{https://liyaowei-stu.github.io/project/BrushEdit}
\thanks{This paper was produced by the IEEE Publication Technology Group. They are in Piscataway, NJ.}
\thanks{Manuscript received April 19, 2021; revised August 16, 2021.}}


\twocolumn[{
\renewcommand\twocolumn[1][]{#1}
\maketitle
\begin{center}
    \centering
    \vspace{-0.9cm}
    \includegraphics[width=0.98\linewidth]{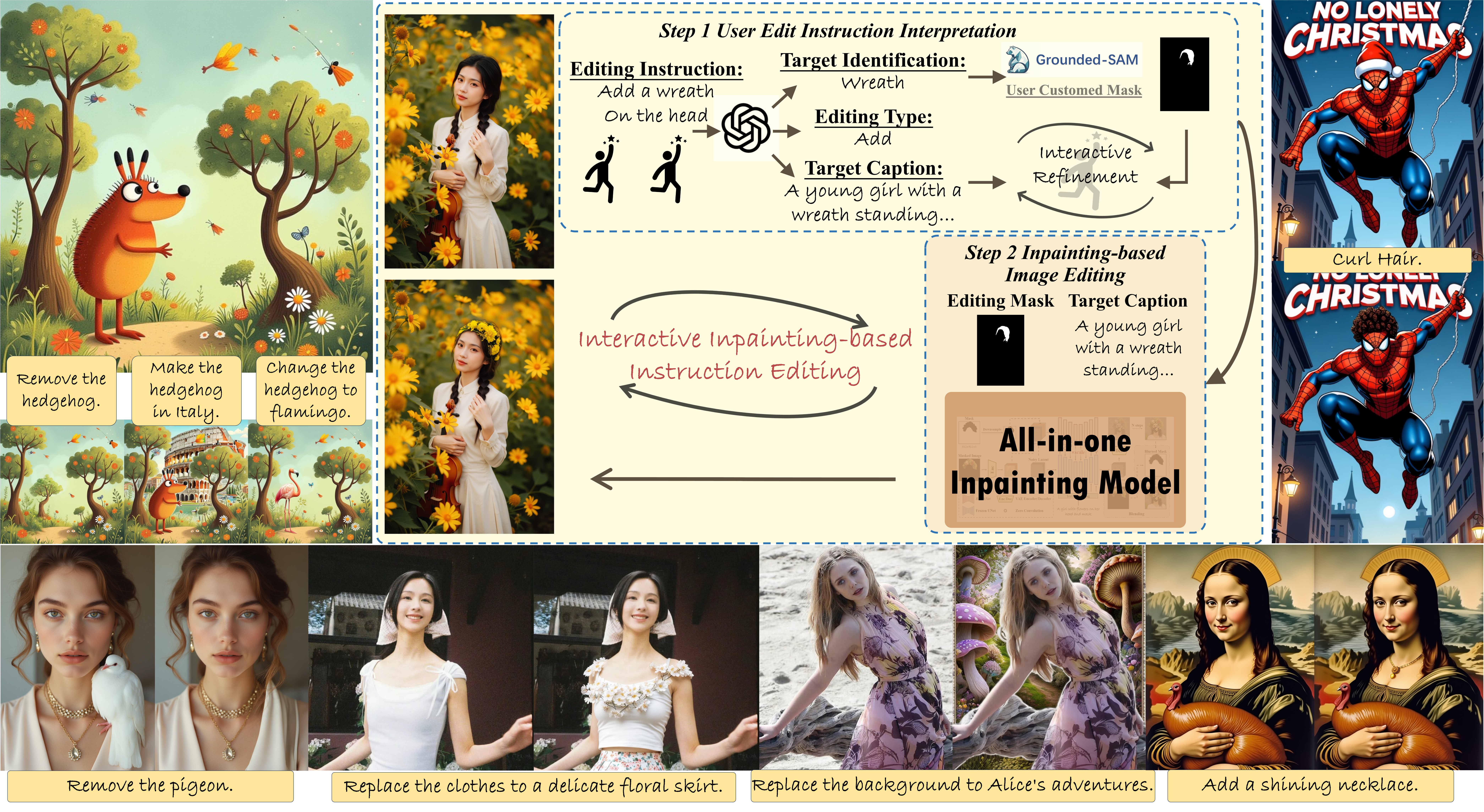}
    \captionof{figure}{\textbf{\OurMethod~is a cutting-edge interactive image editing framework that combines language models and inpainting techniques for seamless edits.} Leveraging pre-trained multimodal language models and BrushNet's dual-branch architecture, users can achieve diverse edits such as adding objects, removing elements, or making structural changes with free-form masks.}
    \label{fig:teaser}
\end{center}
}]

\begin{abstract} 
Image editing has advanced significantly with the development of diffusion models using both inversion-based and instruction-based methods. However, current inversion-based approaches struggle with big modifications (e.g., adding or removing objects) due to the structured nature of inversion noise, which hinders substantial changes. Meanwhile, instruction-based methods often constrain users to black-box operations, limiting direct interaction for specifying editing regions and intensity.  
To address these limitations, we propose BrushEdit, a novel inpainting-based instruction-guided image editing paradigm, which leverages multimodal large language models (MLLMs) and image inpainting models to enable autonomous, user-friendly, and interactive free-form instruction editing.  
Specifically, we devise a system enabling free-form instruction editing by integrating MLLMs and a dual-branch image inpainting model in an agent-cooperative framework to perform editing category classification, main object identification, mask acquisition, and editing area inpainting. Extensive experiments show that our framework effectively combines MLLMs and inpainting models, achieving superior performance across seven metrics including mask region preservation and editing effect coherence.

\end{abstract}


\input{sections/introduction}

\input{sections/related_work}

\input{sections/preliminaries_and_motivation}

\input{sections/method}

\input{sections/experiments}

\input{sections/conclusion}

\bibliographystyle{IEEEtran}
\bibliography{egbib}

\clearpage

\begin{IEEEbiography}[{\includegraphics[width=1in,height=1in,clip,keepaspectratio]{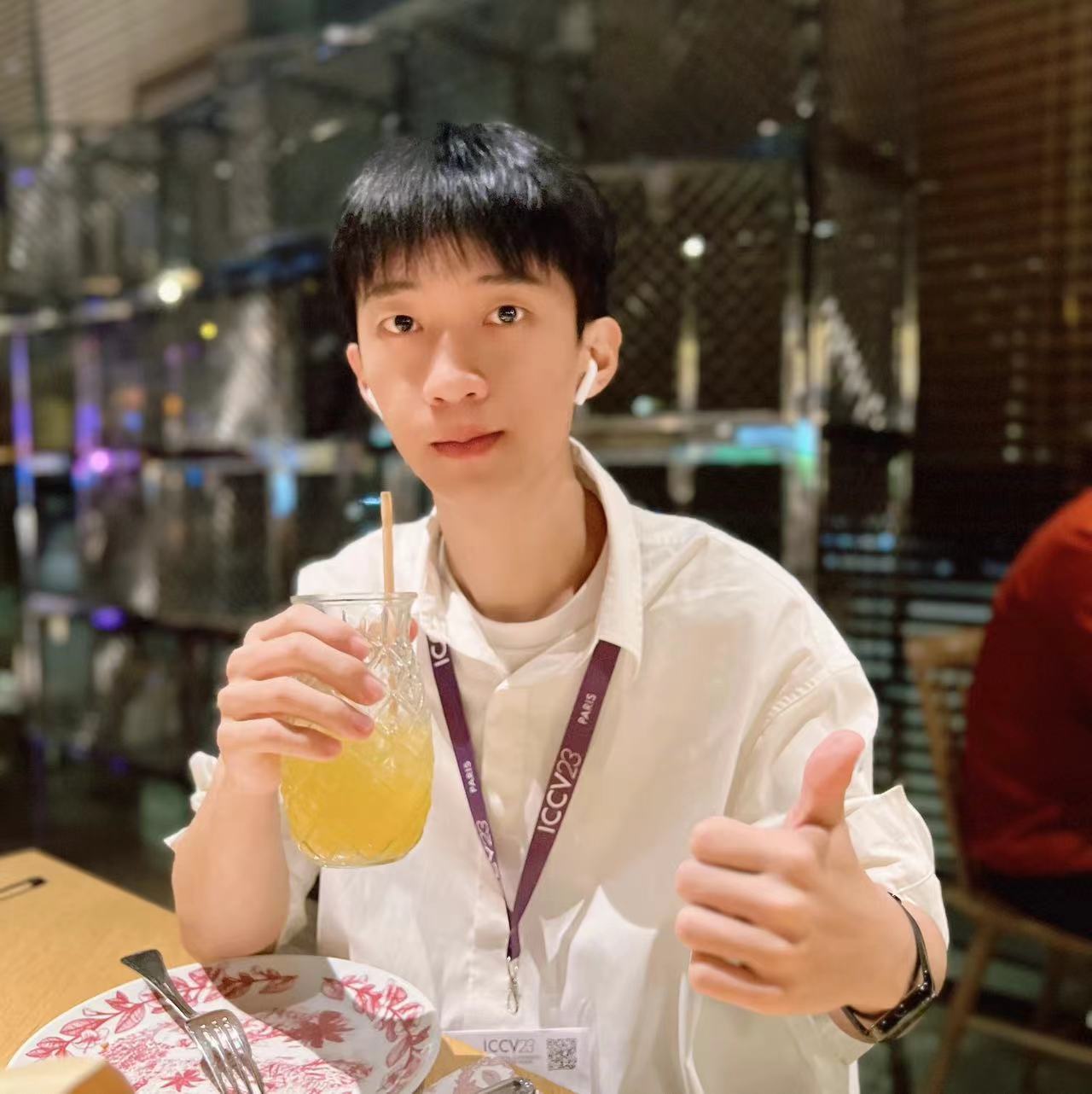}}]{Yaowei Li} is currently pursuing his Ph.D. degree at Peking University. His research interests primarily focus on image/video generation and editing, controllable and interactive media synthesis, and multi-modal processing. He has published several papers in prestigious conferences, including ICCV, ICLR, AAAI, SIGGRAPH, ACL, and EMNLP. He actively contributes to the academic community by serving as a reviewer for leading conferences such as ECCV, NeurIPS, AAAI, CVPR, and ICML.
\end{IEEEbiography}

\begin{IEEEbiography}[{\includegraphics[width=1in,height=1in,clip,keepaspectratio]{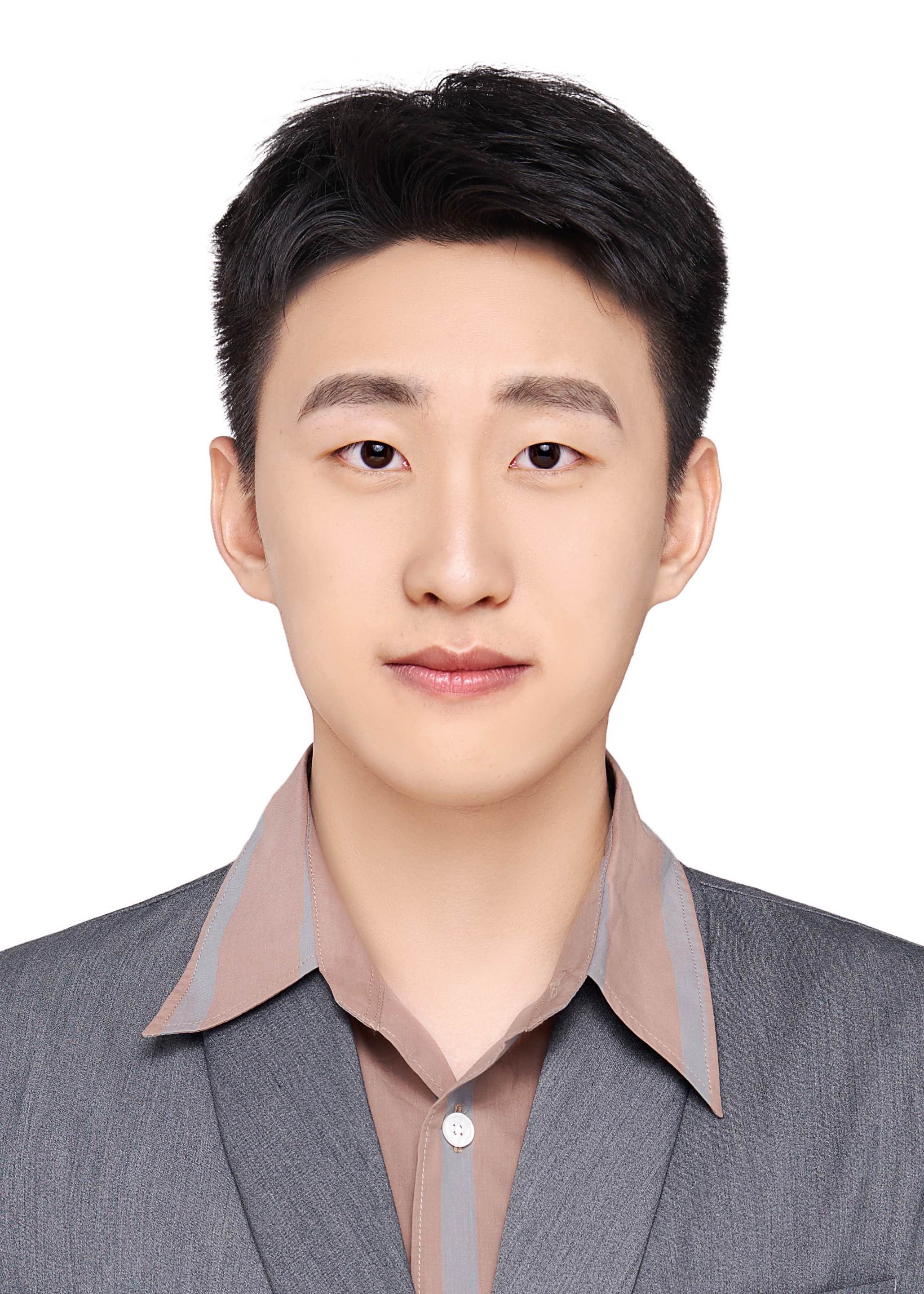}}]{Yuxuan Bian} is currently pursuing a Ph.D. degree at The Chinese University of Hong Kong, under the supervision of Qiang Xu. His research interests include controllable image and video generation, as well as human motion generation.
\end{IEEEbiography}

\begin{IEEEbiography}[{\includegraphics[width=1in,height=1in,clip,keepaspectratio]{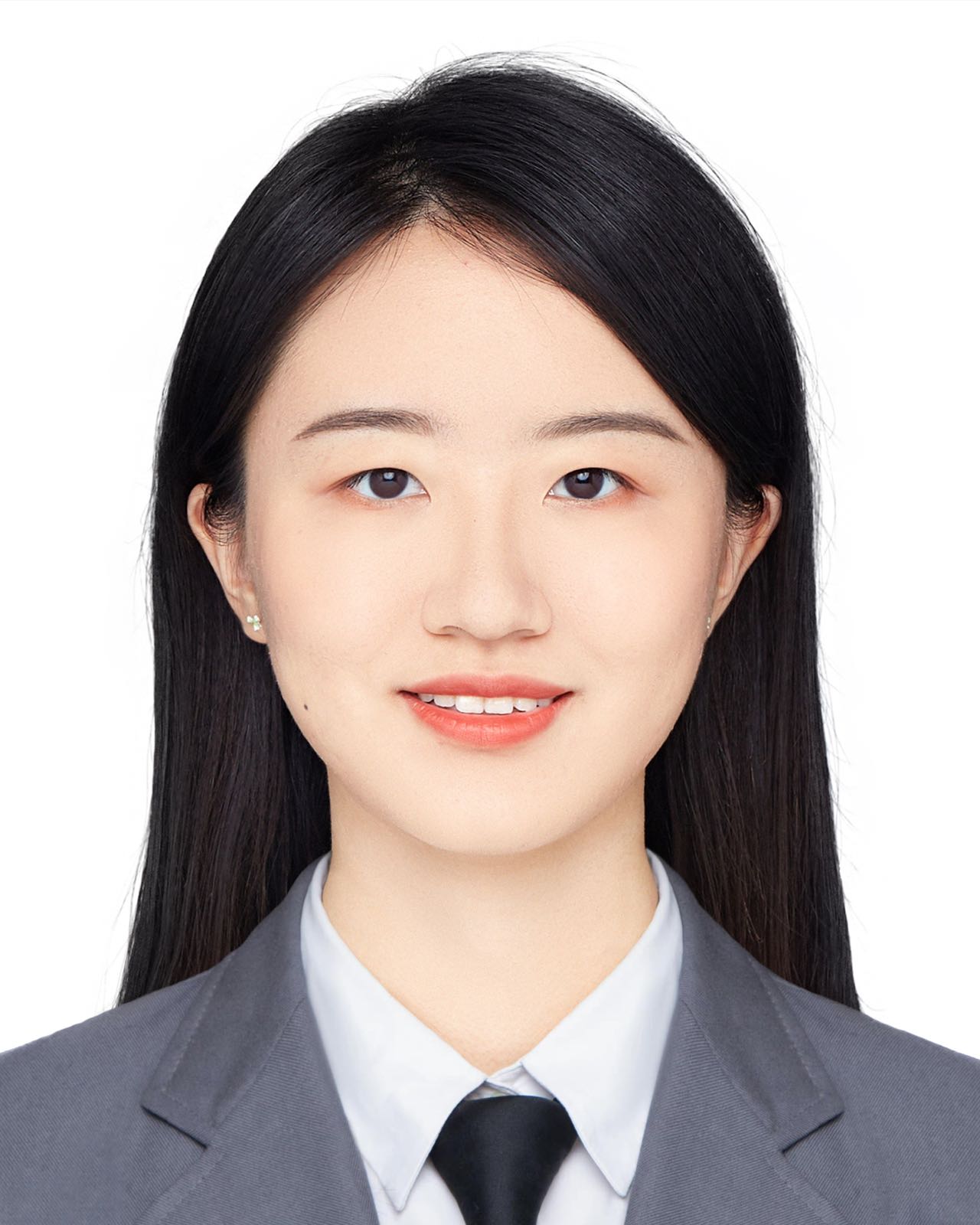}}]{Xuan Ju} is a Ph.D. student at The Chinese University of Hong Kong. Her research focuses on image and video generation, multimodal image/video synthesis, and human-centric visual perception and generation. She has published papers in leading conferences, including CVPR, ECCV, ICCV, NeurIPS, ICLR, and ICML. Additionally, she has organized CVPR workshops and served as a reviewer for top-tier conferences.
\end{IEEEbiography}

\begin{IEEEbiography}[{\includegraphics[width=1in,height=1in,clip,keepaspectratio]{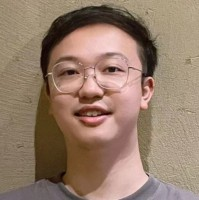}}]{Zhaoyang Zhang} is currently a Senior Research Scientist in ARC Lab, Tencent. He received his Ph.D. degree from The Chinese University of Hong Kong in 2024. His research interests include machine learning, visual generation, and vision-language processing. He has published papers in leading conferences, including CVPR, ICCV, ECCV, NeurIPS, ICML, and ICLR.
\end{IEEEbiography}

\begin{IEEEbiography}[{\includegraphics[width=1in,height=1in,clip,keepaspectratio]{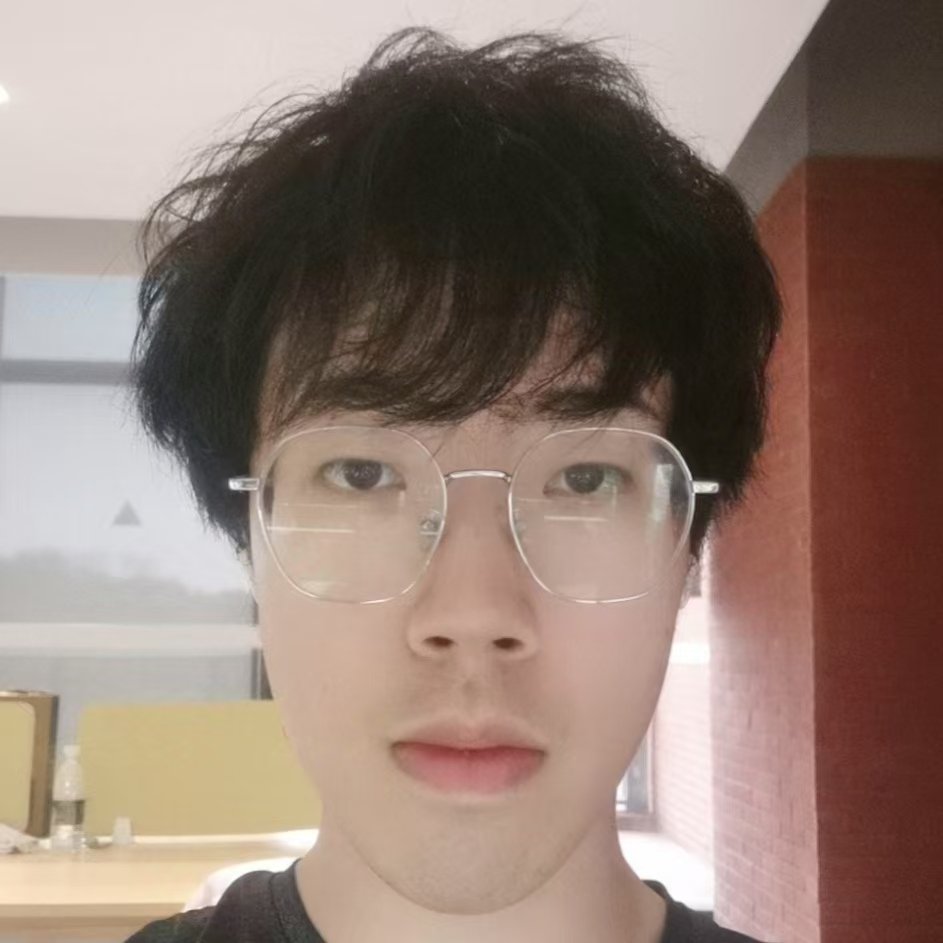}}]{Junhao Zhuang} is a Master student in Computer Technology at Tsinghua University, advised by Professor Chun Yuan. He earned his Bachelor's degree in Computer Science and Technology from the University of Electronic Science and Technology of China. His research focuses on diffusion models, image/video generation, and editing. He has published papers at conferences such as ECCV, ACM MM, and ICASSP.
\end{IEEEbiography}

\begin{IEEEbiography}[{\includegraphics[width=1in,height=1in,clip,keepaspectratio]{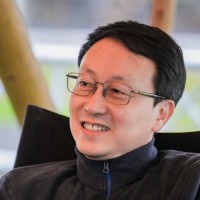}}]{Ying Shan} (Senior Member, IEEE) is a distinguished scientist with Tencent, and the director of the ARC Lab, Tencent PCG. Before joining Tencent, he worked at Microsoft Research as a post-doc researcher, SRI International (Sarnoff Subsidiary) as a senior MTS, and Microsoft Bing Ads as a principal scientist manager. He has published more than 70 papers in top conferences and journals in the areas of computer vision, machine learning, and data mining, served as ACs of CVPR and senior PC of KDD, and holds a number of US/International patents. He is currently leading R\&D efforts in web search, and content AI for a suite of social media and content distribution products.
\end{IEEEbiography}

\begin{IEEEbiography}[{\includegraphics[width=1in,height=1in,clip,keepaspectratio]{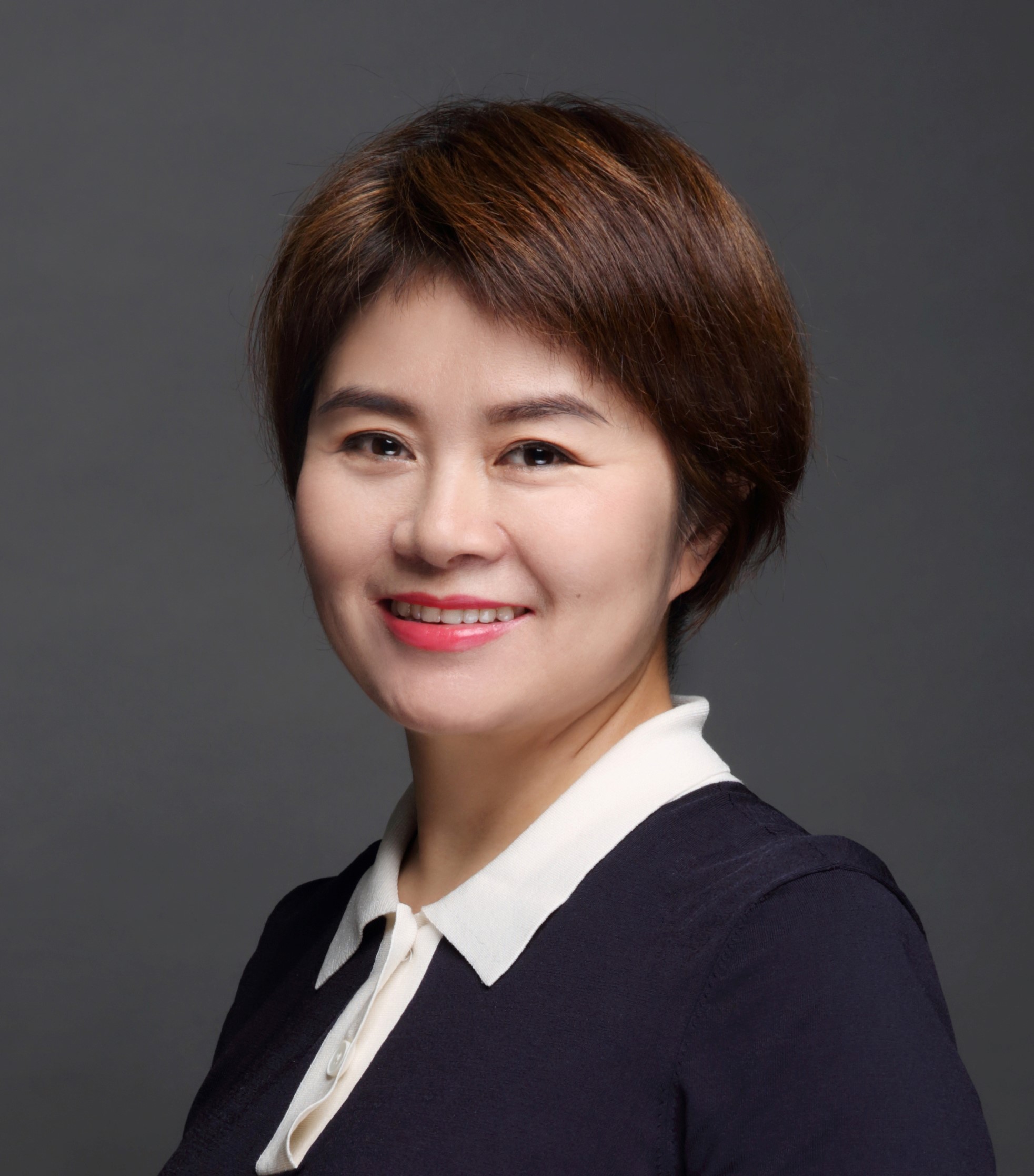}}]{Yuexian Zou} (Senior Member, IEEE) is currently a Full Professor with Peking University and the Director of the Advanced Data and Signal Processing Laboratory in Peking University and serves as the Deputy Director of Shenzhen Association of Artificial Intelligence (SAAI). She was a recipient of the award Leading Figure for Science and Technology by Shenzhen Municipal Government in 2009. She conducted more than 20 research projects including NSFC and 863 projects. She has published more than 280 academic papers in famous journals and flagship conferences, and issued nine invention patents. Her research interests are mainly in intelligent signal and information processing, human-computer voice interaction, video and image processing, and machine learning.
\end{IEEEbiography}

\begin{IEEEbiography}[{\includegraphics[width=1in,height=1in,clip,keepaspectratio]{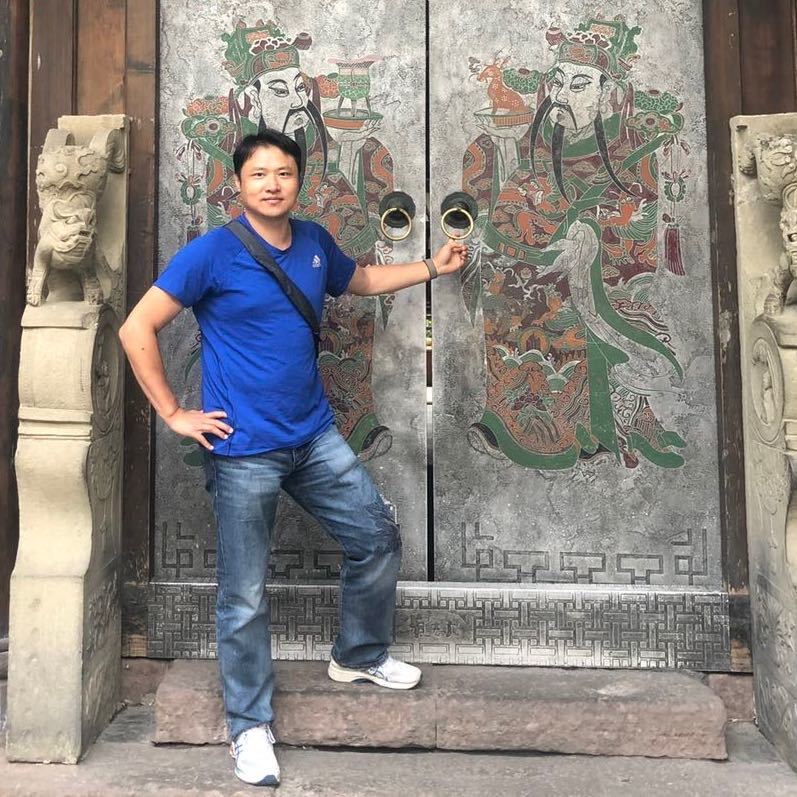}}]{Qiang Xu} is a Professor at The Chinese University of Hong Kong. His research interests include computer vision, large language models, and electronic design automation (EDA). He has published over 200 papers in related fields with more than 11,000 citations, including several best paper awards at prestigious conferences and an ICCAD Ten Year Retrospective Most Influential Paper award.
\end{IEEEbiography}

\vfill

\end{document}

%% file: sections/introduction.tex
\section{Introduction}
\label{sec:introduction}

\IEEEPARstart{T}{he} rapid advancement of diffusion models has significantly propelled text-guided image generation\cite{ho2020denoising,song2020denoising,ju2023humansd,liu2023hyperhuman}, delivering exceptional quality\cite{Rombach_2022_CVPR}, diversity\cite{emu}, and alignment with textual guidance\cite{li2024textcraftor}. However, in image editing tasks—where a target image is generated based on a source image and editing instructions—such progress remains limited due to the difficulty of collecting large amounts of paired data.

To perform image editing based on diffusion generation models, previous methods primarily focus on two strategies:
\textbf{(1) Inversion-based Editing:} This approach leverages the structural information of noised latent derived from inversion to preserve content in non-edited regions, while manipulating the latent in edited regions to achieve the desired modifications \cite{hertz2022prompt,cao2023masactrl,meng2022sdedit,ju2023direct}. Although this method effectively maintains the overall image structure, it is often time-consuming due to multiple diffusion sampling processes. Additionally, the implicit inverse condition significantly limits editability, making large-scale edits (e.g., background replacement) and structural changes (e.g., adding or removing objects) challenging \cite{infedit}. Furthermore, these methods typically require users to provide precise and high-quality source and target captions to leverage the conditional generation model's priors for preserving backgrounds and altering foregrounds. However, in practical scenarios, users often prefer to achieve target area modifications with simple editing instructions.
\textbf{(2) Instruction-based Editing:} This strategy involves collecting paired ``source image-instruction-target image" data and fine-tuning diffusion models for editing tasks \cite{brooks2023instructpix2pix,instructdiffusion,smartedit,mgie}. Due to the difficulty of obtaining manually edited paired data, training datasets are often generated using multimodal large language models (MLLMs) and inversion-based image editing methods (e.g., Prompt-to-Prompt \cite{hertz2022prompt} and Masactrl \cite{cao2023masactrl}). However, the low success rate and unstable quality of these training-free methods \cite{ju2023direct} result in noisy and unreliable datasets, leading to suboptimal performance of the trained models. Additionally, these methods often use a black-box editing process, preventing users from interactively controlling and refining edits \cite{liu2024magicquill}.

\begin{figure*}[htbp]
    \centering
    \includegraphics[width=0.98\linewidth]{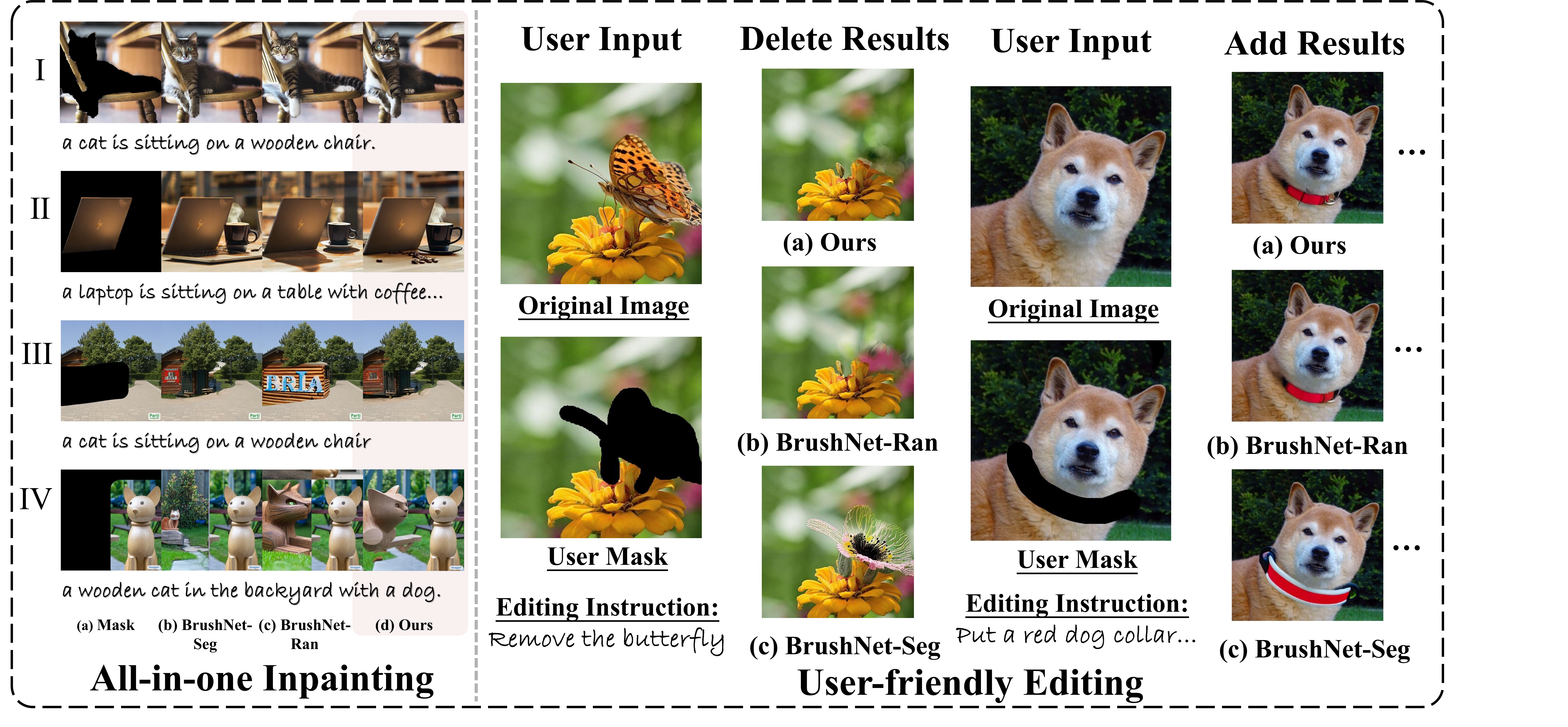}
    \caption{\textbf{\OurMethod~can achieve all-in-one inpainting for arbitrary mask shapes without requiring separate model training for each mask type.} This flexibility in handling arbitrary shapes also enhances user-driven editing, as user-provided masks often combine segmentation-based structural details with random mask noise. By supporting arbitrary mask shapes, \OurMethod~avoids the artifacts introduced by the random-mask version of BrushNet-Ran and the edge inconsistencies caused by the segmentation-mask version BrushNet-Seg's strong reliance on boundary shapes.}
    \label{fig:compare}
\end{figure*}

Given these limitations, we pose the question: Can we develop \textbf{another editing paradigm} that overcomes the challenges of inference efficiency, scalable data curation, editability, and controllability?
The remarkable image-text understanding capabilities of Multimodal Large Language Models (MLLMs) \cite{chen2023internvl,chen2024far,achiam2023gpt,yang2024qwen2}, combined with the outstanding background preservation and text-aligned foreground generation abilities of Image Inpainting models \cite{ju2024brushnet,powerpaint}, inspires us to propose \textbf{\textit{BrushEdit}}. \textit{BrushEdit} is an \textbf{agent-based, free-form, interactive framework} for\textbf{ inpainting-driven} image editing with instruction guidance that highlights the untapped potential in combining language understanding and image generation capabilities to enable free-form, high-quality interactive natural language-based instruction image editing.
This framework requires users to input only natural language editing instructions and supports efficient, arbitrary-round interactive editing, allowing for adjustments in editing types and intensity.

Our approach consists of four main steps:
(i) Editing category classification: determine the type of editing required.
(ii) Identification of the primary editing object: Identify the main object to be edited.
(iii) Acquisition of the editing mask and target Caption: Generate the editing mask and corresponding target caption.
(iv) Image inpainting: Perform the actual image editing.
Steps (i) to (iii) utilize pre-trained MLLMs \cite{achiam2023gpt,yang2024qwen2} and detection models \cite{groundedsam} to ascertain the editing type, target object, editing masks, and target caption. Step (iv) involves image editing using the dual-branch inpainting model \textbf{BrushNet}, as detailed in our previous conference paper. This model inpaints the target areas based on the target caption and editing masks, leveraging the generative potential and background preservation capabilities of inpainting models.
This framework enables steps (i) to (iii) to extract and summarize instructional information via MLLMs, providing clear intermediate interactive guidance for subsequent diffusion models. Meanwhile, step (iv) maximizes the inpainting models' ability to preserve the background and generate foreground content as instructed. 
Meanwhile, users can interactively modify intermediate control information~(e.g., editing mask or the caption of the edited image) during steps (i) to (iv) and iteratively execute these steps as many times as needed until a satisfactory editing result is achieved.
The result is a user-friendly, free-form, multi-turn interactive instruction editing system.

Moreover, we found that BrushNet's original strategy of training separately on segmentation-based masks and random masks greatly limits its practical applicability. This is because these masks differ significantly from user-drawn masks, resulting in suboptimal performance. User-drawn masks often resemble segmentation masks in terms of object edge shapes but also contain noise and irregularities similar to random masks. To overcome this limitation, we refined, merged, and expanded the original BrushData. This allowed us to train an all-in-one inpainting model capable of handling arbitrary mask shapes, thereby facilitating versatile image editing and inpainting, as illustrated in Fig. \ref{fig:compare}.

We present a comprehensive evaluation of \OurMethod through both qualitative and quantitative analyses. We demonstrate that our system significantly enhances image editing quality and efficiency compared to existing methods. It excels particularly in aligning with edit instructions and maintaining background fidelity, thereby validating the effectiveness of our unified inpainting-driven, instruction-guided editing paradigm.

In summary, we extend our conference version\cite{ju2024brushnet} by introducing several novel contributions:
\begin{enumerate}
    \item We introduce BrushEdit, an advanced iteration of the previous BrushNet model. BrushEdit extends the capabilities of controllable image generation by pioneering an inpainting-based image editing approach. This unified model supports instruction-guided image editing and inpainting, offering a user-friendly, free-form, multi-turn interactive editing experience.
    \item By integrating with existing pre-trained multimodal large language models and vision understanding models, BrushEdit significantly improves language comprehension and controllable image generation without necessitating additional training process.
    \item We expand BrushNet into a versatile image inpainting framework that can accommodate arbitrary mask shapes. This eliminates the need for separate models for different types of mask configurations and enhances its adaptability to real-world user masks.
\end{enumerate}

%% file: sections/related_work.tex
\section{Related Work}
\label{sec:related_work}
\subsection{Image Editing}

\input{tables/related}

Image editing involves modifying object shapes, colors, poses, materials, and adding or removing objects\cite{huang2024editingsurvey}. Recent advancements in diffusion models\cite{ho2020denoising,song2020denoising} have notably improved visual generation tasks, outperforming GAN-based models\cite{liu2021pd,zheng2022image,zhao2021large} in image editing. To enable controlled and guided editing, various methods leverage modalities like text instructions\cite{brooks2023instructpix2pix,instructdiffusion,emu}, masks\cite{smartedit,powerpaint,singh2024smartmask}, layouts\cite{hertz2022prompt,cao2023masactrl,epstein2023diffusion}, segmentation maps\cite{matsunaga2022fine,yang2024imagebrush}, and point-dragging interfaces\cite{shi2023dragdiffusion,pan2023drag}. However, these methods often struggle with large structural edits due to noisy latent inversion's overwhelming structural information or rely on scarce high-quality “source image-target image-editing instruction” pairs. Additionally, they usually require users to operate in a black-box manner, demanding precise inputs like masks, text, or layouts, limiting their usability for content creators. These challenges impede the development of a free-form, interactive natural language editing system.

Many Multi-modal Large Language Model~(MLLM)-based methods leverage advanced vision and language understanding capabilities for image editing\cite{smartedit,mgie,liu2024magicquill,nguyen2024flexedit,wang2024genartist}. MGIE refines instruction-based editing by generating more detailed and expressive prompts. SmartEdit enhances the comprehension and reasoning of complex instructions. FlexEdit integrates MLLMs to process image content, masks, and textual inputs. GenArtist employs an MLLM agent to decompose complex tasks, guide tool selection, and systematically execute image editing, generation, and self-correction with iterative verification. However, these methods often involve costly MLLM fine-tuning, are limited to single-turn black-box editing, or face both challenges.

The recent MagicQuill\cite{liu2024magicquill} enables fine-grained control over shape and color at the regional level using scribbles and colors, leveraging a fine-tuned MLLM to infer editing options from user input. While it provides precise interactive control, it requires labor-intensive strokes to define regions and incurs significant training costs to fine-tune MLLMs. In contrast, our method relies solely on natural language instructions (e.g., "remove the rose from the dog's mouth" or "convert the dumplings on the plate to sushi") and integrates MLLMs, detection models, and our dual-branch inpainting mode in a training-free, agent-cooperative framework. And our framework also supports multi-round refinement, users can iteratively adjust the generated editing mask and target image caption to achieve multi-round interaction. 
As summarized in Tab.~\ref{tab:compare_other_methods}, our \OurMethod~overcomes the limitations of current editing methods through an instruction-based, multi-turn interactive, and plug-and-play design, enabling flexible preservation of unmasked regions and establishing itself as a versatile editing solution.

\begin{figure*}[htbp]
    \centering
    \includegraphics[width=1.\linewidth]{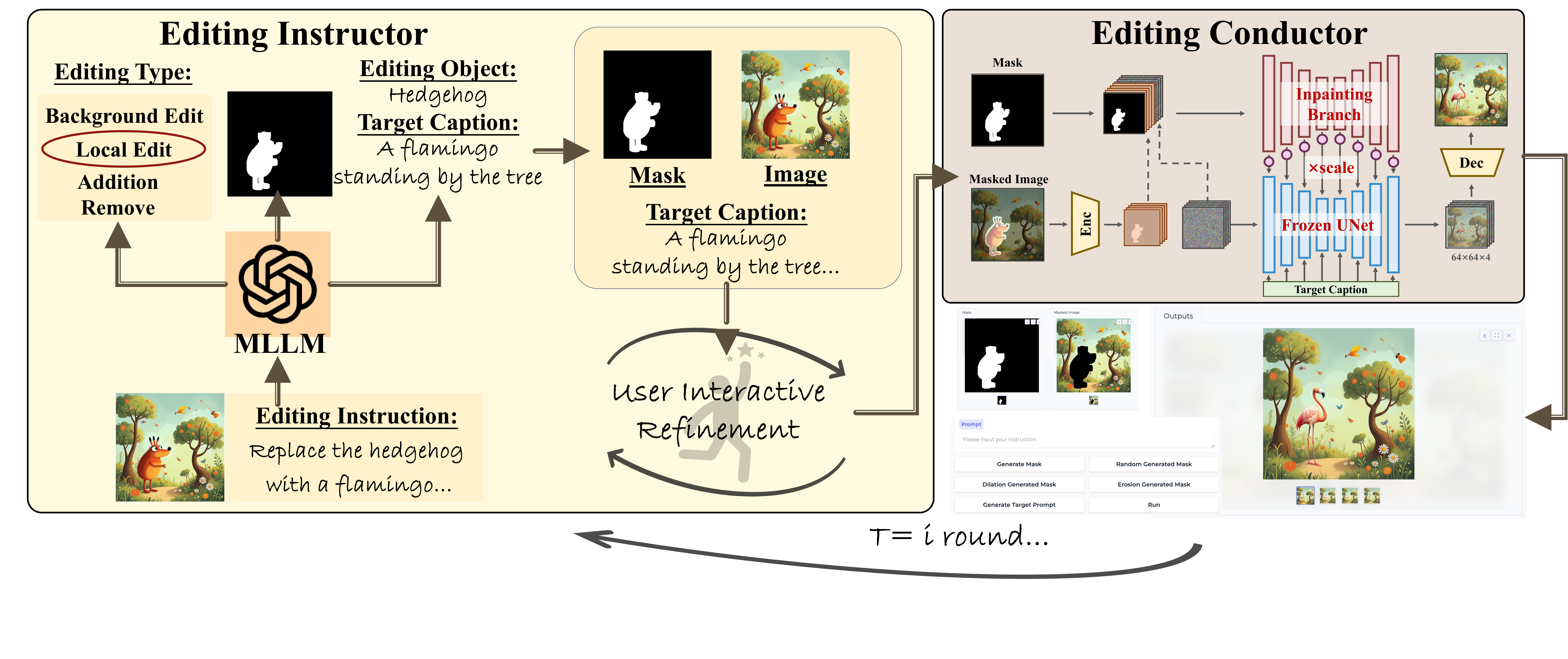}
    \vspace{-40pt}
    \caption{\textbf{Model overview.} Our model outputs an inpainted image given the mask and masked image input. Firstly, we downsample the mask to accommodate the size of the latent, and input the masked image to the VAE encoder to align the distribution of latent space. Then, noisy latent, masked image latent, and downsampled mask are concatenated as the input of \OurMethod. The feature extracted from \OurMethod~is added to pretrained UNet layer by layer after a zero convolution block\cite{zhang2023adding}. After denoising, the generated image and masked image are blended with a blurred mask.}
    \label{fig:model}
\end{figure*}

\subsection{Image Inpainting}
Image inpainting remains a key challenge in computer vision, focusing on reconstructing masked regions with realistic and coherent content\cite{quan2024deep,xu2023review}. Traditional methods\cite{bertalmio2000image,criminisi2004region} and early Variational Auto-Encoder (VAE)\cite{zheng2019pluralistic,peng2021generating} or Generative Adversarial Network (GAN)\cite{liu2021pd,zhao2021large,zheng2022image} approaches often depend on hand-crafted features, leading to limited results.

Recently, diffusion-based models\cite{lugmayr2022repaint,avrahami2022blended,avrahami2023blended,xie2023smartbrush,Kandinsky_2,liu2023image} have gained traction for their superior generation quality, precise control, and diverse outputs\cite{ho2020denoising,ho2022classifier,Rombach_2022_CVPR}. Early diffusion approaches for text-guided inpainting\cite{lugmayr2022repaint,avrahami2022blended,avrahami2023blended,liu2023image,zhang2023coherent,corneanu2024latentpaint,yang2023magicremover}, such as Blended Latent Diffusion, modify denoising by sampling masked regions using pre-trained models while preserving unmasked areas from input images. Despite their popularity in tools like Diffusers\cite{von-platen-etal-2022-diffusers}, these methods perform well on simple tasks but falter with complex masks, content, or prompts, often yielding inconsistent outputs due to limited contextual understanding of mask boundaries and surrounding regions.
To overcome these shortcomings, recent works\cite{xie2023smartbrush,Rombach_2022_CVPR,zhuang2023task,xie2023dreaminpainter,wang2023imagen,ReplaceAnything,yu2023inpaint,yang2023uni} have fine-tuned base models for enhanced content- and shape-awareness. For example, SmartBrush\cite{xie2023smartbrush} integrates object-mask predictions for better sampling, while Stable Diffusion Inpainting\cite{Rombach_2022_CVPR} processes masks, masked images, and noisy latents through the UNet architecture for optimized inpainting. Moreover, HD-Painter\cite{manukyan2023hd} and PowerPaint\cite{zhuang2023task} improve these models for higher quality and multi-task functionality.

However, many methods struggle to generalize inpainting capabilities to arbitrary pre-trained models. One prominent effort is fine-tuning ControlNet\cite{zhang2023adding} on inpainting pairs, but its design remains limited in perceptual understanding, leading to suboptimal results. As summarized in Tab.~\ref{tab:compare_other_methods}, our \OurMethod~addressed these issues with a content-aware, shape-aware, and plug-and-play design, allowing flexible preservation of unmasked regions. Building on this, \OurMethod~unifies training across random and segmentation masks, enabling a single model to handle arbitrary masks seamlessly, advancing its role as a versatile inpainting solution.

%% file: tables/related.tex
\begin{table}[htbp]
    \centering
    \caption{\textbf{Comparison of \OurMethod~with Previous Image Editing/Inpainting Methods.} Note that we only list commonly used text-guided diffusion methods in this table. }
    \scalebox{0.72}{
    \renewcommand\arraystretch{0.8}
\setlength{\tabcolsep}{0.7mm}{
    \begin{tabular}{ccccc}
\toprule
Editing Model                       & Plug-and-Play  & Flexible-Scale & Multi-turn Interactive & Instruction Editing\\ 
\midrule
Prompt2Prompt\cite{hertz2022prompt}        &   \checkmark   &   \checkmark    &        &    \\

MasaCtrl\cite{cao2023masactrl}           &   \checkmark   &   \checkmark    &         &   \\
MagicQuill\cite{liu2024magicquill}           &   \checkmark   &   \checkmark    &     \checkmark    &   \\
InstructPix2Pix\cite{brooks2023instructpix2pix}           &      &       &         &  \checkmark  \\
GenArtist\cite{wang2024genartist}           &   \checkmark   &       &         &  \checkmark  \\
\midrule
\OurMethod            &   \checkmark  &    \checkmark &  \checkmark  &  \checkmark \\
\bottomrule
\toprule
Inpainting Model                       & Plug-and-Play  & Flexible-Scale & Content-Aware & Shape-Aware \\ 
\midrule
Blended Diffusion\cite{avrahami2022blended, avrahami2023blended}           &  \checkmark    &       &        &       \\
SmartBrush\cite{xie2023smartbrush}           &      &       &         &  \checkmark  \\
SD Inpainting\cite{Rombach_2022_CVPR} &               &      &  \checkmark  & \checkmark \\
PowerPaint\cite{zhuang2023task}           &      &       &   \checkmark     &   \checkmark    \\
HD-Painter\cite{manukyan2023hd}           &      &       &   \checkmark     &   \checkmark    \\
ReplaceAnything\cite{ReplaceAnything}             &               &      &  \checkmark  &  \checkmark \\ 
Imagen\cite{wang2023imagen}             &               &      &  \checkmark  &  \checkmark \\ 
ControlNet-Inpainting\cite{zhang2023adding}             &       \checkmark        &  \checkmark    &   \checkmark   &  \\ \midrule
\OurMethod            &   \checkmark  &    \checkmark &  \checkmark  &  \checkmark \\ 
\bottomrule
\end{tabular}}}
    \label{tab:compare_other_methods}
\end{table}

%% file: sections/preliminaries_and_motivation.tex
\section{Preliminaries and Motivation}
\label{sec:preliminaries_and_motivation}

In this section, we will first introduce diffusion models in Sec.~\ref{sec:diffusion_models}. Then, Sec.~\ref{sec:previous_inpainting_models} would review previous inpainting techniques based on sampling strategy modification and special training. Finally, the motivation is outlined in Section~\ref{sec:motivation}.

\subsection{Diffusion Models}
\label{sec:diffusion_models}

Diffusion models include a forward process that adds Gaussian noise $\epsilon$ to convert clean sample $z_0$ to noise sample $z_T$, 
and a backward process that iteratively performs denoising from $z_T$ to $z_0$, where $\epsilon \sim \mathcal{N} \left( 0,1 \right)$, and $T$ represents the total number of timesteps. The forward process can be formulated as follows:

\begin{equation}
 z_t=\sqrt{\alpha _t}z_{0}+\sqrt{1-\alpha _t}\epsilon
 \label{eq:forward}
\end{equation}

$z_t$ is the noised feature at step $t$ with $t\sim \left[ 1,T \right]$, and $\alpha$ is a hyper-parameter.

In the backward process, given input noise $z_T$ sampled from a random Gaussian distribution, learnable network $\epsilon_{\theta}$ estimates noise at each step $t$ conditioned on $C$. After $T$ progressively refining iterations, $z_{0}$ is derived as the output sample:

\begin{equation}
\begin{aligned}
    z_{t-1}= &  \\
    & \hspace{-20pt} \frac{\sqrt{\alpha _{t-1}}}{\sqrt{\alpha _t}}z_t+\sqrt{\alpha _{t-1}}\left( \sqrt{\frac{1}{\alpha _{t-1}}-1}-\sqrt{\frac{1}{\alpha _t}-1} \right) \epsilon _{\theta}\left( z_t,t,C \right) 
 \label{eq:ddim_sample}
\end{aligned}
\end{equation}

The training of diffusion models revolves around optimizing the denoiser network $\epsilon_{\theta}$ to conduct denoising with condition $C$, guided by the objective:

\begin{equation}
\underset{\theta}{\min}E_{z_0,\epsilon \sim \mathcal{N} \left( 0,I \right) ,t\sim U\left( 1,T \right)}\left\| \epsilon -\epsilon _{\theta}\left(z_t, t, C\right) \right\| 
 \label{eq:train_objective}
\end{equation}

\subsection{Image Inpainting Models} 
\label{sec:previous_inpainting_models}

Previous image inpainting approaches can be broadly categorized into \textbf{Sampling Strategy Modification} and \textbf{Dedicated Inpainting Models}.

\textbf{Sampling Strategy Modification.} These methods perform inpainting by iteratively blending masked images with generated content. A representative example is Blended Latent Diffusion (BLD)\cite{avrahami2023blended}, the default inpainting technique in popular diffusion-based libraries (\textit{e.g.}, Diffusers\cite{von-platen-etal-2022-diffusers}). Given a binary mask $m$ and a masked image $x_0^{masked}$, BLD extracts the latent representation $z_0^{masked}$ using a VAE. The mask $m$ is resized to $m^{resized}$ to match the latent dimensions. During inpainting, Gaussian noise is added to $z_0^{masked}$ over $T$ steps, producing $z_t^{masked}$, where $t \sim \left[ 1, T \right]$. The denoising starts from $z_T^{masked}$, with each sampling step (eq.~\ref{eq:ddim_sample}) followed by:

\begin{equation}
z_{t-1} \gets z_{t-1}\cdot\left(1-m^{resized}\right)+z_{t-1}^{masked}\cdot m^{resized}
 \label{eq:blending}
\end{equation}

Despite its simplicity, Sampling Strategy Modification often struggles to preserve unmasked regions and align generated content. These shortcomings stem from:  
(1) inaccuracies introduced by resizing the mask, which hinder proper blending of noisy latents, and  
(2) the diffusion model's limited contextual understanding of mask boundaries and unmasked regions.

\textbf{Dedicated Inpainting Models.}
To enhance performance, these methods fine-tune base models by adding the mask and masked image as additional UNet input channels, creating architectures specialized for inpainting. While they surpass BLD in generation quality, they face several challenges:  
(1) They merge noisy latents, masked image latents, and masks at the UNet's initial convolution layer, where text embeddings globally affect all features, making it difficult for deeper layers to focus on masked image details.  
(2) Simultaneously handling conditional inputs and generation tasks increases the UNet's computational load.  
(3) Extensive fine-tuning is required for different diffusion backbones, leading to high computational costs and limited adaptability to custom diffusion models.

\subsection{Image Editing Models} 
\label{sec:previous_editing_models}

Recent image editing methods can fall into two types:

\paragraph{Inversion Methods}
These approaches\cite{hertz2022prompt,cao2023masactrl,kawar2023imagic,valevski2022unitune,meng2022sdedit, avrahami2022blended} achieve editing by manipulating the latents obtained through inversion. First, they generate edit-friendly noisy latents using various inversion techniques, followed by three paradigms for preserving background regions while modifying target areas:  
(1) \textbf{Attention Integration}: They\cite{hertz2022prompt, han2023improving, parmar2023zero, cao2023masactrl, tumanyan2023plug, zhang2023real, shi2023dragdiffusion} fuse attention maps linking text and image between the source and editing diffusion branches.  
(2) \textbf{Target Embedding}: They\cite{kawar2023imagic, cheng2023general, wu2023uncovering, brack2023sega, tsaban2023ledits, valevski2022unitune, dong2023prompt,cyclediffusion1,cyclediffusion2} manage to embed the editing information from the target branch and integrate it into the source diffusion branch.  
(3) \textbf{Latent Integration}: These methods\cite{meng2022sdedit, avrahami2022blended, avrahami2023blended, couairon2022diffedit, zhang2023sine, shi2023dragdiffusion,joseph2023iterative} try to directly inject editing instructions via noisy latent features from the target diffusion branch into the source diffusion branch.  
Although these methods are computationally efficient and achieve competitive zero-shot or few-shot performance, they are often limited in the diversity of supported edits (e.g., typically restricted to object interaction or attribute modification) due to simplistic generation controls. Additionally, the structural prominence in inversion latents often leads to failures when handling significant structural changes, such as object addition/removal or background replacement.

\paragraph{End-to-end Methods}
These methods\cite{brooks2023instructpix2pix, kim2022diffusionclip, nichol2021glide,geng2023instructdiffusion} train end-to-end diffusion models for image editing, leveraging various ground-truth or pseudo-paired editing datasets. They support a broader range of edits and avoid the significant speed drawbacks of inversion methods, completing edits in a single forward pass. However, their performance is often constrained by the limited availability of ground-truth editing pairs, necessitating pseudo-pair generation via inversion methods, which hinders their upper-bound performance. Furthermore, these end-to-end models lack support for interactive, multi-round editing, preventing content creators from iterative refining or enhancing edits, thus reducing their practicality.

\subsection{Motivation}
\label{sec:motivation}

Based on the analysis in Section~\ref{sec:previous_inpainting_models}, a more effective inpainting architecture could incorporate an additional branch dedicated to processing masked images, enabling the backbone to recognize mask boundaries and the corresponding background without requiring modifications or retraining. 
Similarly, as discussed in Section~\ref{sec:previous_editing_models}, there is an urgent need for a free-form, interactive natural language instruction editing model. Leveraging the exceptional multimodal understanding of MLLMs, such a model can efficiently identify the editing type, target objects, and regions to edit, as well as generate annotations for the desired output. With the support of image inpainting models, precise edits within the target masked regions can then be achieved. Moreover, this process can be iteratively refined, allowing users to create transparently and iteratively.

%% file: sections/method.tex
\section{Method}
\label{sec:method}

An overview of \OurMethod~is shown in Fig.~\ref{fig:model}.  
Our framework integrates MLLMs with a dual-branch image inpainting model via agent collaboration, enabling free-form, multi-turn interactive instruction editing. Specifically, a pre-trained MLLM, acting as the Editing Instructor, interprets user instructions to identify editing types, locate target objects, retrieve detection results for the editing region, and generate textual descriptions of the edited image. Guided by this information, the inpainting model, serving as the Editing Conductor, fills the masked region based on the target text caption. This iterative process allows users to modify or refine intermediate control inputs at any stage, supporting flexible and interactive instruction-based editing.

\subsection{Editing Instructor}

In \OurMethod, we use an MLLM as an editing instructor to interpret users' free-form editing instructions, categorize them into predefined types (addition, removal, local edit, background edit), identify target objects, and utilize a pre-trained detection model to find the relevant editing mask. Finally, the edited image caption is generated. In the next stage, this information is packaged and sent to the editing system to complete the task using an image inpainting approach.

The formal process is as follows: Given the editing instruction $\mathcal{T}_{Ins}$ and source image $\mathcal{I}_{src}$, we first use a pre-trained MLLM \( \phi_{MLLM} \) to identify the user's editing type $\mathcal{K}$ and the corresponding target object $\mathcal{O}$ The MLLM then calls a pre-trained detection model \( \phi_{D} \) to search for the target object mask $\mathcal{M}_{d}$ based on $\mathcal{O}$. After obtaining the mask, the MLLM combines $\mathcal{K}$, $\mathcal{O}$, and $\mathcal{I}_{src}$ to generate the final edited image caption. The source image $\mathcal{I}_{src}$, target mask $\mathcal{M}_{d}$, and the caption are then passed to the next stage, the Editing Conductor, for image-inpainting-based editing.

\subsection{Editing Conductor}

Our Editing Conductor, built on our previous BrushNet, employs a mixed fine-tuning strategy using both random and segmentation masks. This approach enables the inpainting model to handle diverse mask-based inpainting tasks without being restricted by mask types, achieving comparable or superior performance. Specifically, we inject masked image features into a pre-trained diffusion network (e.g., Stable Diffusion 1.5) through an additional control branch. These features include the noisy latent for enhancing semantic coherence by providing information on the current generation process, the masked image latent extracted via VAE to guide semantic consistency between the prompt foreground and the ground truth background, and the mask downsampled via cubic interpolation to explicitly indicate the position and boundaries of the foreground filling region.

To retain masked image features, \OurMethod~employs a duplicate of the pre-trained diffusion model with all attention layers removed. The pre-trained convolutional weights serve as a robust prior for extracting masked image features, while excluding cross-attention layers ensures the branch focuses solely on pure background information. \OurMethod~features are integrated into the frozen diffusion model layer-by-layer, enabling hierarchical, dense per-pixel control. Following ControlNet\cite{zhang2023adding}, zero convolution layers are used to link the frozen model with the trainable \OurMethod, mitigating noise during early training stages.
The feature insertion operation is defined in Eq.~\ref{eq:insertion}:

\begin{equation}
\begin{aligned}
\epsilon _{\theta}\left( z_t,t,C \right) _i= &  \\
& \hspace{-55pt} \epsilon _{\theta}\left( z_t,t,C \right) _i+w\cdot \mathcal{Z} \left( \epsilon _{\theta}^{BrushNet}\left( \left[ z_t,z_{0}^{masked},m^{resized} \right] ,t \right) _i \right) 
  \label{eq:insertion}
\end{aligned}
\end{equation}

, where $\epsilon _{\theta}\left( z_t,t,C \right) _i$ represents the feature of the $i$-th layer in the network $\epsilon _{\theta}$, where $i \in \left[ 1,n \right]$, and $n$ denotes the total number of layers. The same notation is applied to $\epsilon _{\theta}^{BrushNet}$. The network $\epsilon _{\theta}^{BrushNet}$ processes the concatenated noisy latent $z_t$, masked image latent $z_{0}^{masked}$, and downsampled mask $m^{resized}$, where concatenation is represented by $\left[ \cdot \right]$. $\mathcal{Z}$ refers to the zero convolution operation, and $w$ is the preservation scale that adjusts the influence of \OurMethod~on the pretrained diffusion model.

Previous studies have highlighted that downsampling during latent blending can introduce inaccuracies, and the VAE encoding-decoding process has inherent limitations that impair full image reconstruction. To ensure consistent reconstruction of unmasked regions, prior methods have explored various strategies. Some approaches\cite{zhuang2023task,ReplaceAnything} rely on copy-and-paste techniques to directly transfer unmasked regions, but these often result in outputs lacking semantic coherence. Latent blending methods inspired by BLD\cite{avrahami2023blended,Rombach_2022_CVPR} also struggle to retain desired information in unmasked areas effectively. In this work, we propose a simple pixel-space approach that applies mask blurring before copy-and-paste using the blurred mask. Although this may slightly affect accuracy near the mask boundary, the error is nearly imperceptible and significantly improves boundary coherence.

The architecture of \OurMethod~is inherently designed for seamless plug-and-play integration with various pretrained diffusion models, enabling flexible preservation control. Specifically, the flexible capabilities of \OurMethod~include:  
(1) Plug-and-Play Integration: As \OurMethod~does not modify the pretrained diffusion model’s weights, it can be effortlessly integrated with any community fine-tuned models, facilitating easy adoption and experimentation.  
(2) Preservation Scale Adjustment: The preservation scale of the unmasked region can be controlled by incorporating \OurMethod~features into the frozen diffusion model with a weight $w$, which adjusts the influence of \OurMethod~on the level of preservation.  
(3) Blurring and Blending Customization: The preservation scale can be further refined by adjusting the blurring scale and applying blending operations as needed.  
These features provide fine-grained and flexible control over the editing process.

%% file: sections/experiments.tex
\section{Experiments}
\label{sec:experiments}

\subsection{Evaluation Benchmark and Metrics}

\begin{figure*}[htbp]
    \centering
    \includegraphics[width=0.99\linewidth]{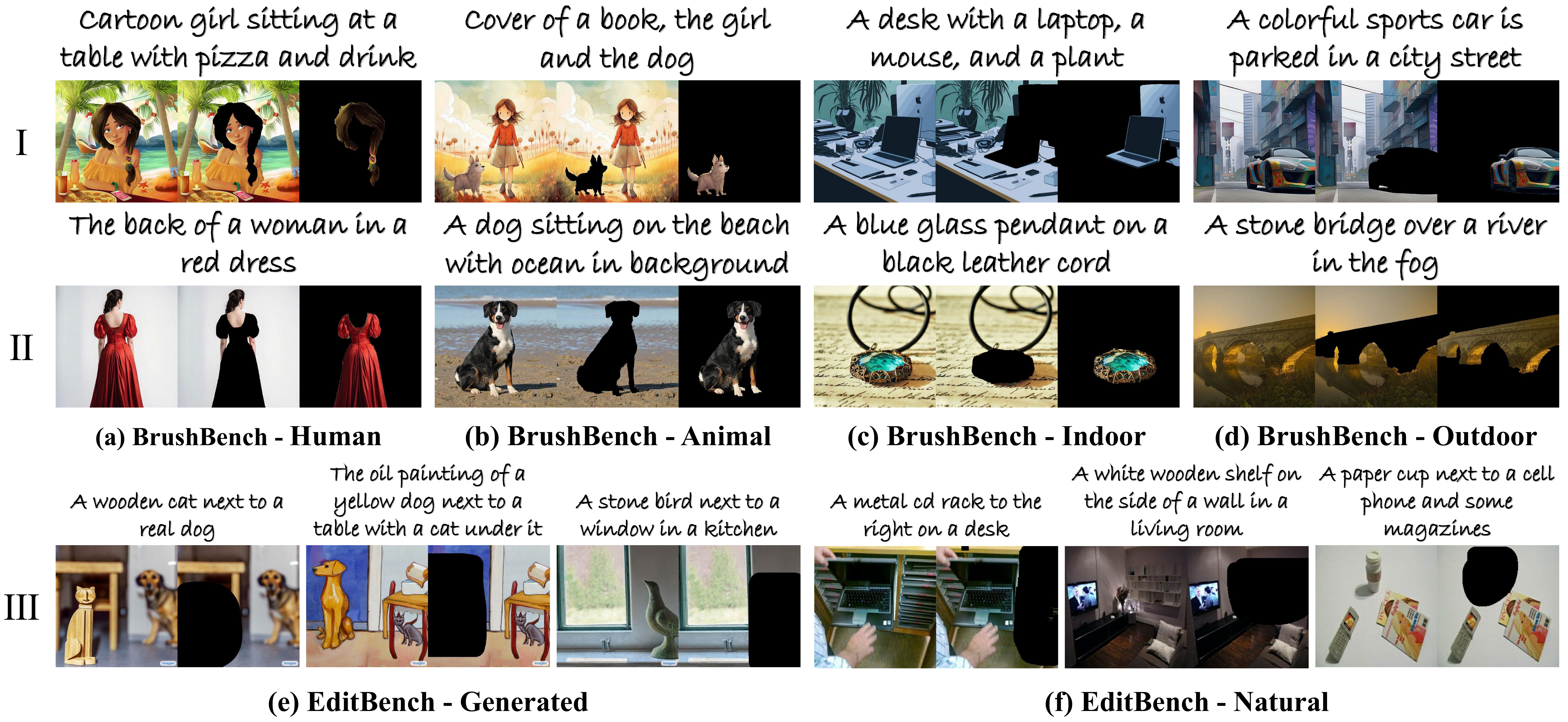}
    \caption{\textbf{Benchmark overview.} \uppercase\expandafter{\romannumeral1} and \uppercase\expandafter{\romannumeral2} separately show natural and artificial images, masks, and caption of \Benchmark. (a) to (d) show images of humans, animals, indoor scenarios, and outdoor scenarios. Each group of images shows the original image, inside-inpainting mask, and outside-inpainting mask, with an image caption on the top. \uppercase\expandafter{\romannumeral3} show image, mask, and caption from EditBench~\cite{wang2023imagen}, with (e) for generated images and (f) for natural images. The images are randomly selected from both benchmarks.}
    \label{fig:benchmark}
\end{figure*}

\paragraph{\textbf{Benchmark}}  
To comprehensively evaluate the performance of \OurMethod, we conducted experiments on both image editing and image inpainting benchmarks:
\begin{itemize}
    \item  \textbf{Image Editing.} We used PIE-Bench~\cite{ju2023direct} (\textbf{P}rompt-based \textbf{I}mage \textbf{E}diting \textbf{Bench}mark) to evaluate \OurMethod and all baselines on image editing tasks. PIE-Bench consists of $700$ images spanning $10$ editing types, evenly distributed between natural and artificial scenes (\emph{e.g.}, paintings) across four categories: animal, human, indoor, and outdoor. Each image includes five annotations: source image prompt, target image prompt, editing instruction, main editing body, and editing mask. 
    \item  \textbf{Image Inpainting.} Extending our prior conference work, we replaced traditional benchmarks~\cite{liu2015faceattributes,huang2018introvae,deng2009imagenet,lin2014microsoft,kuznetsova2020open,yu2015lsun} with \Benchmark for segmentation-based masks and EditBench for random brush masks. These benchmarks span real and generated images across human bodies, animals, and indoor and outdoor scenes. EditBench includes $240$ images with an equal mix of natural and generated content, each annotated with a mask and caption. \Benchmark, shown in Fig.~\ref{fig:benchmark}, contains $600$ images with human-annotated masks and captions, evenly distributed across natural and artificial scenes (\emph{e.g.}, paintings) and covering various categories such as humans, animals, and indoor/outdoor environments.  
\end{itemize}
%
We refined the task into two scenarios for segmentation-based mask inpainting: inside-inpainting and outside-inpainting, enabling detailed performance evaluation across distinct image regions.

\textbf{Notably, \OurMethod~surpasses BrushNet by leveraging unified high-quality inpainting masked images for training, enabling it to handle all mask types.} This establishes \OurMethod as a unified model capable of performing all inpainting and editing benchmark tasks, whereas BrushNet required separate fine-tuning for each mask type.

\paragraph{\textbf{Dataset}} 
Building upon the \TrainingData~proposed in our previous conference version, we integrate two subsets of segmentation masks and random masks, and further extend the data from the Laion-Aesthetic~\cite{schuhmann2022laion} dataset, resulting in \TrainingData-v2. A key difference is that we select images with clean backgrounds and pair them randomly with either segmentation or random masks, effectively creating pairs that simulate deletion-based editing, significantly enhancing our framework's removal capability in image editing. The data expansion process is as follows: We use Grounded-SAM~\cite{ren2024grounded} to annotate open-world masks, then filter them based on confidence scores to retain only those with higher confidence. We also consider mask size and continuity during the filtering.

\paragraph{\textbf{Metrics}} We evaluate five metrics, focusing on unedited/uninpainted region preservation and edited/inpainted region text alignment. Additionally, we conducted extensive user studies to validate the superior performance of \OurMethod~in edit instruction alignment and background fidelity.

\begin{itemize}
    \item[$\bullet$] \textit{Background Fidelity.} We adopt standard metrics, including Peak Signal-to-Noise Ratio (\textbf{PSNR})~\cite{psnr}, Learned Perceptual Image Patch Similarity (\textbf{LPIPS})~\cite{zhang2018unreasonable}, Mean Squared Error (\textbf{MSE})~\cite{mse}, and  Structural Similarity Index Measure  (\textbf{SSIM})~\cite{wang2004image}, to evaluate the consistency between the unmasked regions of the generated and original images.
    
    \item[$\bullet$] \textit{Text Alignment.} We use CLIP Similarity (\textbf{CLIP Sim})~\cite{clipsim} to assess text-image consistency by projecting both into the shared embedding space of the CLIP model~\cite{clip} and measuring the similarity of their representations.
\end{itemize}

\subsection{Implementation Details}

We evaluate various inpainting methods under a consistent setting unless stated otherwise, \emph{i.e.}, using NVIDIA Tesla V100 GPUs and their open-source code with Stable Diffusion v1.5 as the base model, 50 steps, and a guidance scale of 7.5. 
Each method utilizes its recommended hyper-parameters across all images to ensure fairness. 
\OurMethod~ and all ablation models are trained for $430$k steps on 8 NVIDIA Tesla V100 GPUs, requiring approximately 3 days. 
\textbf{Notably, for all image editing (PnPBench) and image inpainting (\Benchmark~and EditBench) tasks, \OurMethod achieves unified image editing and inpainting using a single model trained on \TrainingData-v2.} In contrast, our previous BrushNet required separate training and testing for different mask types. Additional details are available in the provided code.

\subsection{Quantitative Comparison~(Image Editing)}

\begin{figure*}[htbp]
    \centering
    \vspace{-10pt}
    \includegraphics[width=1.\linewidth]{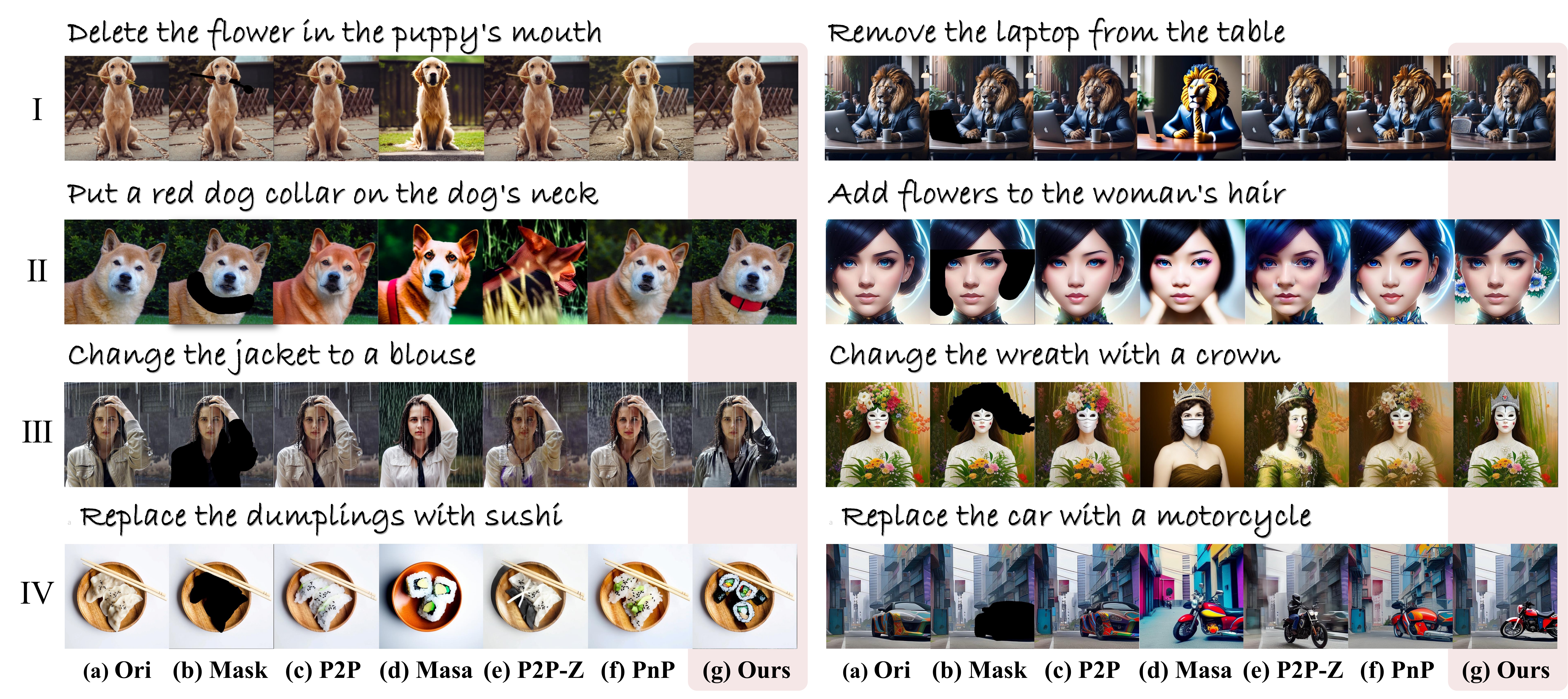}
    \caption{\textbf{Comparison of previous editing methods and \OurMethod~on natural and synthetic images, covering image editing operations such as removing objects~(\uppercase\expandafter{\romannumeral1}), adding objects~(\uppercase\expandafter{\romannumeral2}), modifying attributes~(\uppercase\expandafter{\romannumeral3}), and swapping objects~(\uppercase\expandafter{\romannumeral4}).}}
    \label{fig:experiment_compare_edit}
\end{figure*}

Tab.~\ref{tab:ours_edit_bench} and Tab.~\ref{tab:ours_edit_bench_time} compare the quantitative image editing performance on PnPBench~\cite{ju2023direct}. We evaluate the editing results of previous inversion-based methods, including four inversion techniques—DDIM Inversion~\cite{song2020denoising}, Null-Text Inversion~\cite{mokady2023null}, Negative-Prompt Inversion~\cite{miyake2023negative}, and StyleDiffusion~\cite{li2023stylediffusion}—as well as four editing methods: Prompt-to-Prompt~\cite{hertz2022prompt}, MasaCtrl~\cite{cao2023masactrl}, pix2pix-zero~\cite{parmar2023zero}, and Plug-and-Play~\cite{tumanyan2023plug}.

The results in Tab.~\ref{tab:ours_edit_bench} confirm the superiority of \OurMethod~in preserving unedited regions and ensuring accurate text alignment in edited areas. While inversion-based methods, such as DDIM Inversion (DDIM)~\cite{song2020denoising} and PnP Inversion (PnP)~\cite{ju2023direct}, can achieve high-quality background preservation, they are inherently limited by reconstruction errors that affect background retention. In contrast, \OurMethod~separately models unedited background information through a dedicated branch, while the main network generates the edited region based on the text prompt. Combined with predefined user masks and blending operations, it ensures near-lossless background preservation and semantically coherent edits.

More importantly, our method preserves high-fidelity background information without being affected by the irretrievable structural noise from inversion-based methods. It allows operations, such as adding or removing objects, that are typically impossible with inversion-based editing. Furthermore, since no inversion is required, \OurMethod~only needs a single forward pass to perform the editing operation. As shown in Tab.~\ref{tab:ours_edit_bench_time} , the editing time of \OurMethod~is significantly short, greatly improving the efficiency of image editing.

\input{tables/compare_previous_editing_pnpbench}

\input{tables/compare_infer_time}

\subsection{Qualitative Comparison~(Image Editing)}

The qualitative comparison with previous image editing methods is shown in Fig.~\ref{fig:experiment_compare_edit}. 
We present results on both artificial and natural images across various editing tasks, including deleting objects (\uppercase\expandafter{\romannumeral1}), adding objects (\uppercase\expandafter{\romannumeral2}), modifying objects (\uppercase\expandafter{\romannumeral3}), and swapping objects (\uppercase\expandafter{\romannumeral4}). 
\OurMethod~consistently achieves superior coherence between the edited and unedited regions, excelling in adherence to editing instructions, smoothness at the editing mask boundaries, and overall content consistency. 
Notably, Fig.~\ref{fig:experiment_compare_edit} \uppercase\expandafter{\romannumeral1} and \uppercase\expandafter{\romannumeral2} involve tasks such as deleting a flower or laptop, and adding a collar or earring. 
While previous methods failed to deliver satisfactory results due to persistent structural artifacts caused by inversion noise, 
\OurMethod~successfully performs the intended operations and produces seamless edits that blend harmoniously with the background, owing to its dual-branch decoupled inpainting-based editing paradigm.

\subsection{Quantitative Comparison~(Image Inpainting)}

\input{tables/compare_previous_inpainting_our_bench}

Tab.~\ref{tab:ours_bench} and Tab.~\ref{tab:editbench} present the quantitative comparison on \Benchmark~and EditBench~\cite{wang2023imagen}. We evaluate the inpainting results of the sampling strategy modification method Blended Latent Diffusion~\cite{avrahami2023blended}, dedicated inpainting models Stable Diffusion Inpainting~\cite{Rombach_2022_CVPR}, HD-Painter~\cite{manukyan2023hd}, PowerPaint~\cite{zhuang2023task}, the plug-and-play method ControlNet~\cite{zhang2023adding} trained on inpainting data, and our previous BrushNet\footnote{BrushNet fine-tunes separate models for different mask types, while \OurMethod~uses a unified model and achieves state-of-the-art performance on both segmentation-based \Benchmark~and random-mask-based EditBench.}.

Results confirm \OurMethod's superiority in preserving uninpainted regions and ensuring accurate text alignment in inpainted areas.  
Blended Latent Diffusion~\cite{avrahami2023blended} performs the worst, primarily due to incoherent transitions between masked and unmasked regions, stemming from its disregard for mask boundaries and blending-induced latent space losses.  
HD-Painter~\cite{manukyan2023hd} and PowerPaint~\cite{zhuang2023task}, both based on Stable Diffusion Inpainting~\cite{Rombach_2022_CVPR}, achieve similar results to their base model for inside-inpainting tasks.  
However, their performance deteriorates sharply in outside-inpainting, as they are designed exclusively for inside-inpainting.  
ControlNet~\cite{zhang2023adding}, explicitly trained for inpainting, shares the most comparable experimental setup with ours.  
Nonetheless, its design mismatch with the inpainting task hampers its ability to maintain masked region fidelity and text alignment, requiring integration with Blended Latent Diffusion~\cite{avrahami2023blended} for reasonable results. Even with this combination, it falls short of specialized inpainting models and \OurMethod.
The performance on EditBench aligns closely with that on \Benchmark, both demonstrating \OurMethod's superior results. This suggests that our method performs consistently well across various inpainting tasks, including segmentation, random, inside, and outside inpainting masks.

It is worth noting that, compared to BrushNet, \OurMethod~now surpasses BrushNet in both segmentation-mask-based and random-mask-based benchmarks with a single model, achieving a more general and robust all-in-one inpainting. This improvement is largely attributed to our unified mask types and the richer data distribution in \TrainingData-v2.

\input{tables/compare_previous_general_setting}

\subsection{Qualitative Comparison~(Image Inpainting)}

\begin{figure*}[htbp]
    \centering
    \vspace{-0.2cm}
    \includegraphics[width=0.99\linewidth]{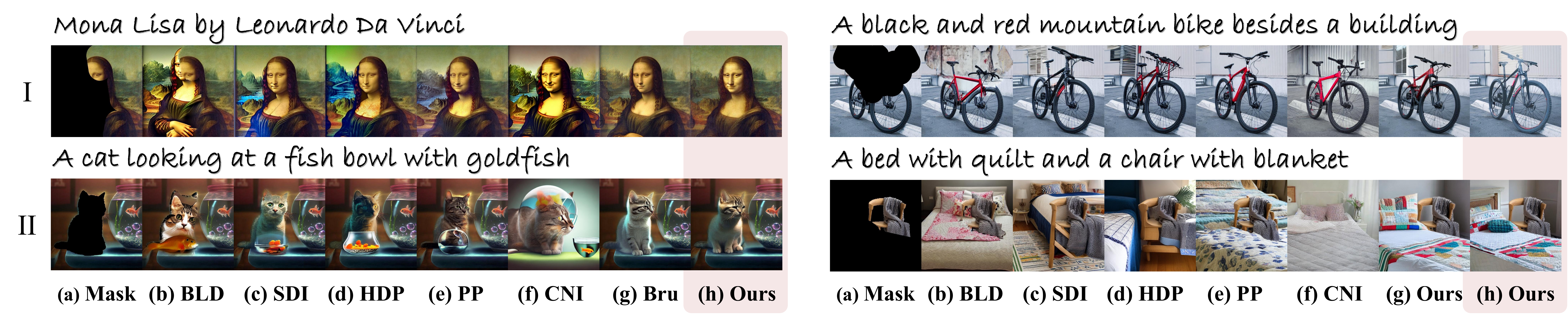}
    \vspace{-10pt}
    \caption{\textbf{Performance comparisons of \OurMethod~and previous image inpainting methods} across various inpainting tasks: (\uppercase\expandafter{\romannumeral1}) Random Mask Inpainting (\uppercase\expandafter{\romannumeral2}) Segmentation Mask Inpainting. Each group of results contains $7$ inpainting methods: (b) Blended Latent Diffusion (BLD)~\cite{avrahami2023blended}, (c) Stable Diffusion Inpainting (SDI)~\cite{Rombach_2022_CVPR}, (d) HD-Painter (HDP)~\cite{manukyan2023hd}, (e) PowerPaint (PP)~\cite{zhuang2023task}, (f) ControlNet-Inpainting (CNI)~\cite{zhang2023adding}, (g) Our Previous BrushNet and (h) Ours.
    }
    \label{fig:experiment_compare_inpaint}
    \vspace{-0.1cm}
\end{figure*}

The qualitative comparison with previous image inpainting methods is shown in Fig.~\ref{fig:experiment_compare_inpaint}.  
We evaluate results on both artificial and natural images across diverse inpainting tasks, including random mask inpainting and segmentation mask inpainting.  
\OurMethod~consistently achieves superior coherence between the generated and unmasked regions in terms of both content and color (\uppercase\expandafter{\romannumeral1}, \uppercase\expandafter{\romannumeral2}).  
Notably, in Fig.~\ref{fig:experiment_compare_inpaint} \uppercase\expandafter{\romannumeral2} (left), the task involves generating both a cat and a goldfish.  
While all prior methods fail to recognize the existing goldfish in the masked image and instead generate an additional fish,  
\OurMethod~accurately integrates background context, enabled by its dual-branch decoupling design.  
Furthermore, \OurMethod~outperforms our previous BrushNet in overall inpainting performance without fine-tuning for specific mask types, achieving comparable or even better results on both random and segmentation-based masks.  

\subsection{Flexible Control Ability}
\label{sec:flexible_control_ability}

\begin{figure*}[htbp]
    \centering
    \includegraphics[width=1.\linewidth]{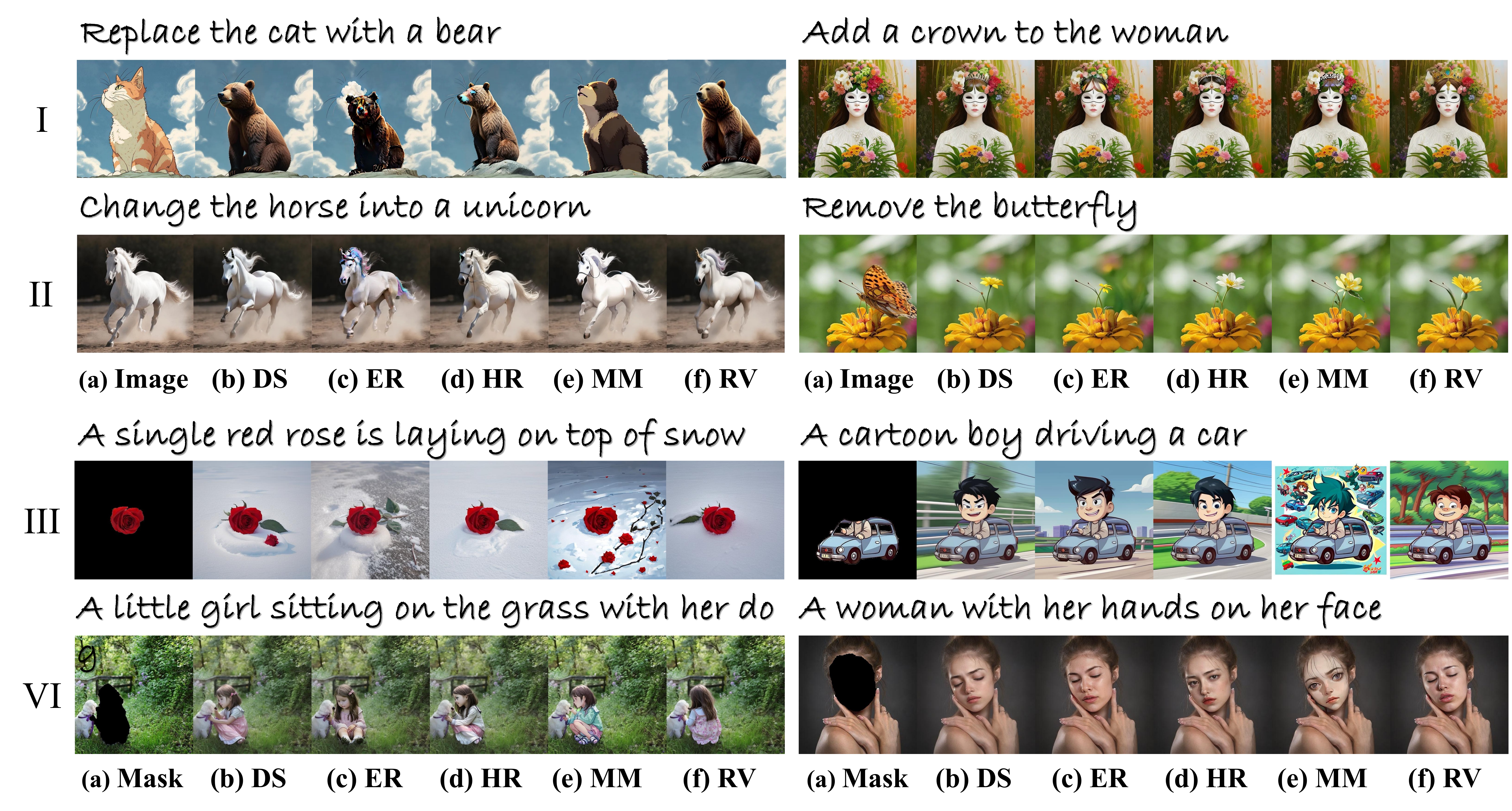}
    \vspace{-0.6cm}
    \caption{\textbf{Integrating \OurMethod~to community fine-tuned diffusion models.} We use five popular community diffusion models fine-tuned from stable diffusion v1.5: DreamShaper (DS)~\cite{dreamshaper}, epiCRealism (ER)~\cite{epiCRealism}, Henmix\_Real (HR)~\cite{HenmixReal}, MeinaMix (MM)~\cite{MeinaMix}, and Realistic Vision (RV)~\cite{RealisticVision}. MM is specifically designed for anime images.}
    \vspace{-0.2cm}
    \label{fig:base_model}
\end{figure*}

\begin{figure*}[htbp]
    \centering
    \includegraphics[width=1.\linewidth]{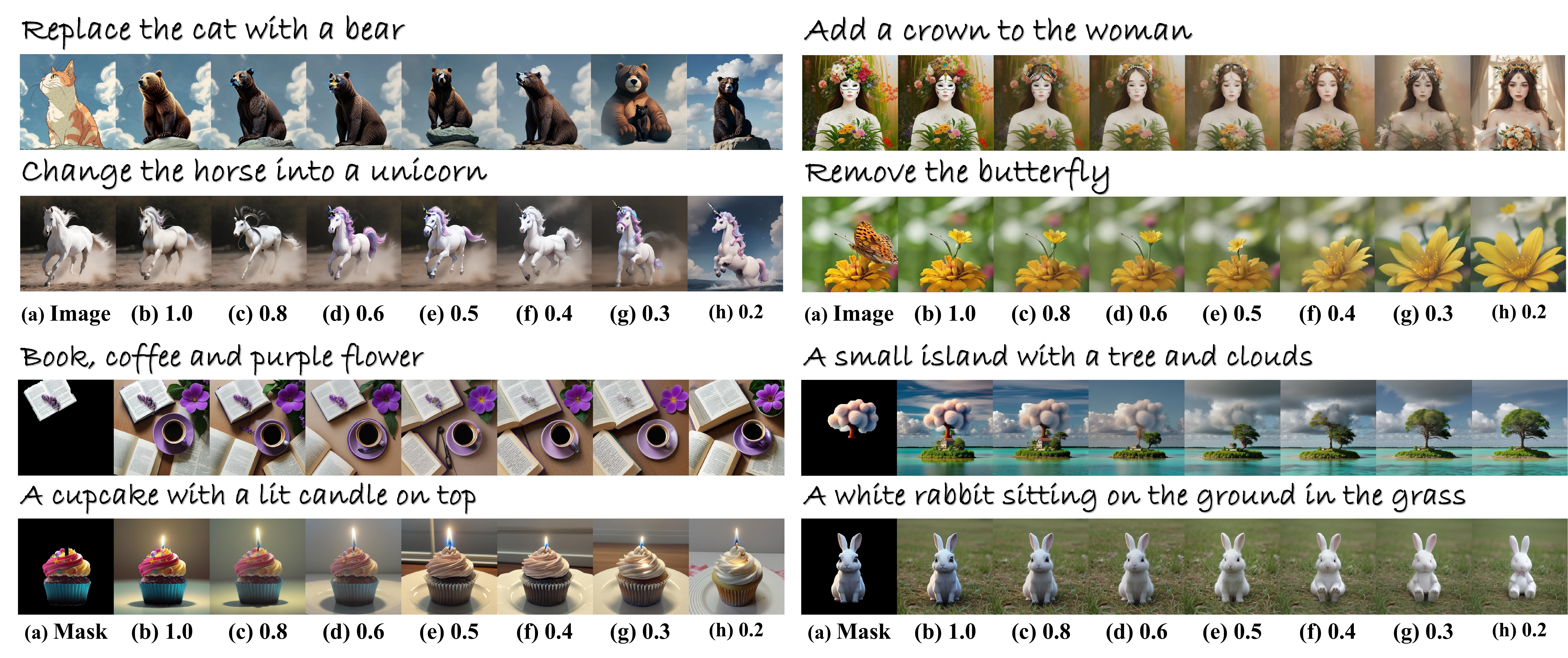}
    \caption{\textbf{Flexible control scale of \OurMethod.} (a) shows the given masked image, (b)-(h) show adding control scale $w$ from $1.0$ to $0.2$. Results show a gradually diminishing controllable ability from precise to rough control.}
    \label{fig:scale}
\end{figure*}

Fig.~\ref{fig:base_model} and Fig.~\ref{fig:scale} demonstrate the flexible control offered by \OurMethod~in two key areas: base diffusion model selection and scale adjustment. This flexibility extends beyond inpainting to image editing, as it is achieved by altering the backbone network's generative prior and branch information injection strength. In Fig.~\ref{fig:base_model}, we show how \OurMethod~can be combined with various community-finetuned diffusion models, enabling users to choose the model that best aligns with their specific editing or inpainting needs. This greatly enhances the practical value of \OurMethod. Fig.~\ref{fig:scale} illustrates the control over \OurMethod's scale parameter, which allows users to adjust the extent of unmasked region protection during editing or inpainting, offering fine-grained control for precise and customizable results.

\subsection{Ablation Study}

\input{tables/ablation_finetune}

We conducted ablation studies to examine the impact of different model designs on image inpainting tasks. Since \OurMethod~ is based on an image inpainting model, the editing task is achieved through inference-only by chaining MLLMs, \OurMethod, and an image detection model as agents. The inpainting capability directly reflects our model's training outcome. Tab.~\ref{tab:ablation_finetune} compares the dual-branch and single-branch designs, while Tab.~\ref{tab:ablation} highlights the ablation study on the additional branch architecture. 

The ablation studies, performed on \Benchmark, average the performance for both inside-inpainting and outside-inpainting. The results in Tab.~\ref{tab:ablation_finetune} show that the dual-branch design significantly outperforms the single-branch design. Moreover, fine-tuning the base diffusion model in the dual-branch setup yields superior results compared to freezing it. However, fine-tuning may limit flexibility and control over the model. 
Considering the trade-off between performance and flexibility, we chose to adopt the frozen dual-branch design for our model. Tab.~\ref{tab:ablation} explains the reasoning behind key design choices: (1) using a VAE encoder instead of randomly initialized convolution layers for processing masked images, (2) incorporating the full UNet feature layer-by-layer into the pre-trained UNet, and (3) removing text cross-attention in \OurMethod to prevent masked image features from being influenced by text.

\input{tables/ablation}

%% file: tables/compare_previous_editing_pnpbench.tex
\begin{table}[!htbp]
\vspace{-0.5cm}
\caption{\textbf{Comparison of \OurMethod~with various editing methods in PnpBench.} For editing methods Prompt-to-Prompt (P2P)\cite{hertz2022prompt}, MasaCtrl\cite{cao2023masactrl}, Pix2Pix-Zero (P2P-Zero)\cite{cao2023masactrl}, and Plug-and-Play (PnP)\cite{tumanyan2023plug}, we evaluate two inversion techniques, DDIM Inversion (DDIM)\cite{song2020denoising} and PnP Inversion (PnP)\cite{ju2023direct}, to establish stronger baselines.  \textcolor{Red}{\textbf{Red}} stands for the best result, \textcolor{Blue}{\textbf{Blue}} stands for the second best result.}

\resizebox{0.48\textwidth}{!}{ 
\small
\centering
\renewcommand\arraystretch{0.6}
\setlength{\tabcolsep}{0.2mm}{
\begin{threeparttable}
{
\begin{tabular}{c|c|cccc|c}
\toprule
\textbf{Inverse}          & \textbf{Editing}          & \textbf{PSNR} $\uparrow$     & \textbf{LPIPS}$_{^{\times 10^3}}$ $\downarrow$  & \textbf{MSE}$_{^{\times 10^4}}$ $\downarrow$     & \textbf{SSIM}$_{^{\times 10^2}}$ $\uparrow$    & \textbf{CLIP Similariy}  $\uparrow$       \\ \midrule
\textbf{DDIM}& \textbf{P2P}          & 17.87  & 208.80  & 219.88  & 71.14       &  {22.44}          \\
\midrule
\textbf{PnP}             & \textbf{P2P}           & {27.22} & {54.55} & {32.86} & {84.76}&  {22.10}         \\
\midrule
\textbf{DDIM}& \textbf{MasaCtrl}            & 22.17  & 106.62  & 86.97  & 79.67        & 21.16          \\
\midrule
\textbf{PnP}& \textbf{MasaCtrl}       & {22.64}  & {87.94}  & {81.09} & {81.33}    & {21.35}         \\
\midrule
\textbf{DDIM}& \textbf{P2P-Zero}  & 20.44  & 172.22 & 144.12 & 74.67 & 20.54 \\
\midrule
\textbf{PnP}& \textbf{P2P-Zero} & {21.53} & {138.98} & {127.32} & {77.05}  & {21.05} \\
\midrule
\textbf{DDIM}& \textbf{PnP}  & 22.28  & 113.46 & 83.64 & 79.05 & \textcolor{Blue}{\textbf{22.55}} \\
\midrule
\textbf{PnP}& \textbf{PnP}  & {22.46} & {106.06}& {80.45} & {79.68}  & \textcolor{Red}{\textbf{22.62}} \\
\toprule
\multicolumn{2}{c|}{\OurMethod}   & \textcolor{Red}{\textbf{32.16}} & \textcolor{Red}{\textbf{17.22}}& \textcolor{Red}{\textbf{8.43}} & \textcolor{Red}{\textbf{97.08}} & {22.44} \\
\bottomrule
\end{tabular}}
\end{threeparttable}}
} 

\label{tab:ours_edit_bench}
\vspace{-0.5cm}
\end{table}

%% file: tables/compare_infer_time.tex
\begin{table}[!htbp]

\caption{\textbf{Comparison of inference time between our inpainting-based \OurMethod~ and other inversion-based methods, including Negative-Prompt Inversion (NP), Edit Friendly Inversion (EF),  AIDI\cite{pan2023effective}, EDICT, Null-Text Inversion (NT), and Style Diffusion added with Prompt-to-Prompt. \OurMethod~achieves better editing results with far less inference time than all inversion-based methods.}}

\resizebox{0.48\textwidth}{!}{  
\small
\centering
  \renewcommand\arraystretch{0.4}
\setlength{\tabcolsep}{0.4mm}{
\begin{threeparttable}
{
\begin{tabular}{c|ccccccc}
\toprule
\textbf{Methods}          & \textbf{\OurMethod}  & \textbf{NP}    & \textbf{EF}  &  \textbf{AIDI}    &  \textbf{EDICT}   &   \textbf{NT}  &  \textbf{Style Diffusion}     \\ 
\midrule
\textbf{Inference Time (s)}   & \textcolor{Red}{\textbf{3.57}} & \textcolor{Blue}{\textbf{18.22}}& 19.10 & 35.41  & 35.48 & 148.48 & 382.98 \\
\bottomrule
\end{tabular}}
    \end{threeparttable}}
    \label{tab:ours_edit_bench_time}
}
\vspace{-10pt}
\end{table}

%% file: tables/compare_previous_inpainting_our_bench.tex
\begin{table*}[htbp]
\centering
\scriptsize
\caption{\textbf{Quantitative comparisons between \OurMethod~and other diffusion-based inpainting models in \Benchmark}: Blended Latent Diffusion (BLD)\cite{avrahami2023blended}, Stable Diffusion Inpainting (SDI)\cite{Rombach_2022_CVPR}, HD-Painter (HDP)\cite{manukyan2023hd}, PowerPaint (PP)\cite{zhuang2023task}, ControlNet-Inpainting (CNI)\cite{zhang2023adding}, and our previous Segmentation-based BrushNet-Seg\cite{ju2024brushnet}. The table shows metrics on background fidelity and text alignment (Text Align) for both inside- and outside-inpainting. All models use Stable Diffusion V1.5 as the base model. \textcolor{Red}{\textbf{Red}} indicates the best result, while \textcolor{Blue}{\textbf{Blue}} indicates the second-best result. }
\scalebox{0.95}{
\setlength{\tabcolsep}{0.9mm}{
\begin{threeparttable}
{
\begin{tabular}{cl|cccc|c||cl|cccc|c}
\toprule
\toprule
\multicolumn{7}{c||}{\textbf{Inside-inpainting}} & \multicolumn{7}{c}{\textbf{Outside-inpainting}}\\
\midrule
\multicolumn{2}{c|}{\textbf{Metrics}}  & \multicolumn{4}{c|}{\textbf{Masked Background Fidelity}} & \textbf{Text Align} & 
\multicolumn{2}{c|}{\textbf{Metrics}}  & \multicolumn{4}{c|}{\textbf{Masked Background Fidelity}} & \textbf{Text Align}\\
\midrule
\multicolumn{2}{c|}{\textbf{Models}}    & \textbf{PSNR}$\uparrow$     & \textbf{MSE}$_{^{\times 10^3}}$$\downarrow$  & \textbf{LPIPS}$_{^{\times 10^3}}$$\downarrow$ & \textbf{SSIM}$_{^{\times 10^3}}$$\uparrow$ & \textbf{CLIP Sim}$\uparrow$ &
\multicolumn{2}{c|}{\textbf{Models}}    & \textbf{PSNR}$\uparrow$     & \textbf{MSE}$_{^{\times 10^3}}$$\downarrow$  & \textbf{LPIPS}$_{^{\times 10^3}}$$\downarrow$ & \textbf{SSIM}$_{^{\times 10^3}}$$\uparrow$ & \textbf{CLIP Sim}$\uparrow$\\ 
\midrule
& \textbf{\xspace BLD (1)} &21.33&9.76&49.26&74.58&26.15 &
& \textbf{\xspace BLD (1)} &15.85&35.86&21.40&77.40&26.73\\
& \textbf{\xspace SDI (2)} &21.52&13.87&48.39&89.07&26.17 &
& \textbf{\xspace SDI (2)} &18.04&19.87&15.13&91.42&27.21\\
& \textbf{\xspace HDP (3)} &22.61&9.95&43.50&89.03&26.37 &
& \textbf{\xspace HDP (3)} &18.03&22.99&15.22&90.48&26.96\\
& \textbf{\xspace PP (4)} &21.43&32.73&48.43&86.39&\textcolor{Red}{\textbf{26.48}} &
& \textbf{\xspace PP (4)} &18.04&31.78&15.13&90.11&26.72\\ 
& \textbf{\xspace CNI (5)} &12.39&78.78&243.62&65.25&\textcolor{Blue}{\textbf{26.47}} &
& \textbf{\xspace CNI (5)} &11.91&83.03&58.16&66.80&\textcolor{Red}{\textbf{27.29}}\\ 
& \textbf{\xspace CNI* (5)} &22.73&24.58&43.49&91.53&26.22 &
& \textbf{\xspace CNI* (5)} &17.50&37.72&19.95&94.87&26.92\\ 
& \textbf{\xspace BrushNet-Seg}* &\textcolor{Blue}{\textbf{31.94}}&\textcolor{Blue}{\textbf{0.80}}&\textcolor{Red}{\textbf{18.67}}&\textcolor{Blue}{\textbf{96.55}}&26.39 &
& \textbf{\xspace BrushNet-Seg}* &\textcolor{Red}{\textbf{27.82}}&\textcolor{Red}{\textbf{2.25}}&\textcolor{Red}{\textbf{4.63}}&\textcolor{Blue}{\textbf{98.95}}&\textcolor{Blue}{\textbf{27.22}}\\ 
& \textbf{\xspace \OurMethod}* &\textcolor{Red}{\textbf{31.98}}&\textcolor{Red}{\textbf{0.79}}&\textcolor{Blue}{\textbf{18.92}}&\textcolor{Red}{\textbf{96.68}}&26.24 &
& \textbf{\xspace \OurMethod}* &\textcolor{Blue}{\textbf{27.65}}&\textcolor{Blue}{\textbf{2.30}}&\textcolor{Blue}{\textbf{4.90}}&\textcolor{Red}{\textbf{98.97}}&\textcolor{Red}{\textbf{27.29}}\\ 
\bottomrule
\bottomrule
\end{tabular}
\begin{tablenotes}
\footnotesize
\item[*] with blending operation
\end{tablenotes}
}
\end{threeparttable}
}
}
\label{tab:ours_bench}
\vspace{-0.5cm}
\end{table*}

%% file: tables/compare_previous_general_setting.tex
\begin{table}[htbp]
\centering
\scriptsize
\caption{\textbf{Quantitative comparisons among \OurMethod~and other diffusion-based inpainting models, Random-mask-based BrushNet-Ran in EditBench}. A detailed explanation of compared methods and metrics can be found in the caption of Tab.~\ref{tab:ours_bench}. \textcolor{Red}{\textbf{Red}} stands for the best result, \textcolor{Blue}{\textbf{Blue}} stands for the second best result.}

\scalebox{0.95}{
\renewcommand\arraystretch{1.2}
\setlength{\tabcolsep}{0.9mm}{
\begin{threeparttable}
{
\begin{tabular}{l|cccc|c}
\toprule
\toprule
\textbf{Metrics}  & \multicolumn{4}{c|}{\textbf{Masked Background Fidelity}} & \textbf{Text Align} \\
\midrule
\textbf{Models}    & \textbf{PSNR}$\uparrow$     & \textbf{MSE}$_{^{\times 10^3}}$$\downarrow$  & \textbf{LPIPS}$_{^{\times 10^3}}$$\downarrow$ & \textbf{SSIM}$_{^{\times 10^3}}$$\uparrow$ & \textbf{CLIP Sim}$\uparrow$     \\ \midrule
\textbf{BLD}\cite{avrahami2023blended} & 20.89&10.93&31.90&85.09&28.62 \\
\textbf{SDI}\cite{Rombach_2022_CVPR} & 23.25&6.94&24.30&90.13&28.00 \\
\textbf{HDP}\cite{manukyan2023hd} & 23.07&\textcolor{Blue}{\textbf{6.70}}&24.32&92.56&28.34 \\
\textbf{PP}\cite{zhuang2023task} & 23.34&20.12&24.12&91.49&27.80 \\
\textbf{CNI}\cite{zhang2023adding} & 12.71&69.42&159.71&79.16&28.16  \\ 
\textbf{CNI}*\cite{zhang2023adding} & 22.61&35.93&26.14&94.05&27.74   \\ 
\textbf{BrushNet-Ran}* & \textcolor{Red}{\textbf{33.66}}&\textcolor{Red}{\textbf{0.63}}&\textcolor{Blue}{\textbf{10.12}}&\textcolor{Blue}{\textbf{98.13}}&\textcolor{Blue}{\textbf{28.87}}\\ 
\textbf{\OurMethod}* & \textcolor{Blue}{\textbf{32.97}}&\textcolor{Blue}{\textbf{0.70}}&\textcolor{Red}{\textbf{7.24}}&\textcolor{Red}{\textbf{98.60}}&\textcolor{Red}{\textbf{29.62}} \\ 
     \bottomrule
     \bottomrule
\end{tabular}
\begin{tablenotes}
\footnotesize
\item[*] with blending operation
\end{tablenotes}
}
\end{threeparttable}
}
}
\label{tab:editbench}
\vspace{-0.5cm}
\end{table}

%% file: tables/ablation_finetune.tex
\begin{table}[htbp]
\centering
\scriptsize
\caption{\textbf{Ablation on dual-branch design.} Stable Diffusion Inpainting (SDI) use single-branch design, where the entire UNet is fine-tuned. We conducted an ablation analysis by training a dual-branch model with two variations: one with the base UNet fine-tuned, and another with the base UNet forzened. Results demonstrate the superior performance achieved by adopting the dual-branch design. \textcolor{Red}{\textbf{Red}} is the best result.}
\scalebox{0.85}{
\setlength{\tabcolsep}{0.9mm}{
\begin{threeparttable}
{
\begin{tabular}{c|ccc|ccc|c}
\toprule
\toprule
\multicolumn{1}{c|}{\textbf{Metrics}}  & \multicolumn{3}{c|}{\textbf{Image Quality}} & \multicolumn{3}{c|}{\textbf{Masked Region Preservation}} & \textbf{Text Align} \\
\midrule
\textbf{Model} & \textbf{IR}$_{^{\times 10}}$$\uparrow$   & \textbf{HPS}$_{^{\times 10^2}}$$\uparrow$  & \textbf{AS}$\uparrow$   & \textbf{PSNR}$\uparrow$     & \textbf{MSE}$_{^{\times 10^2}}$$\downarrow$  & \textbf{LPIPS}$_{^{\times 10^3}}$$\downarrow$      & \textbf{CLIP Sim}$\uparrow$     \\ \midrule
SDI & 11.00 & 27.53 & 6.53 & 19.78 & 16.87 & 31.76 & 26.69 \\ 
w/o fine-tune & 11.59 & 27.71 & 6.59 & 19.86 & 16.09 & 31.68 & 26.91 \\ 
w/ fine-tune & \textcolor{Red}{\textbf{11.63}} & \textcolor{Red}{\textbf{27.73}} & \textcolor{Red}{\textbf{6.60}} & \textcolor{Red}{\textbf{20.13}} & \textcolor{Red}{\textbf{15.84}} & \textcolor{Red}{\textbf{31.57}} & \textcolor{Red}{\textbf{26.93}} \\ 

     \bottomrule
     \bottomrule
\end{tabular}
}
\end{threeparttable}
}
}
\label{tab:ablation_finetune}
\vspace{-0.3cm}
\end{table}

%% file: tables/ablation.tex
\begin{table}[htbp]
\centering
\scriptsize
\caption{\textbf{Ablation on model architecture.} 
We ablate on the following components: the image encoder (Enc), selected from a random initialized convolution (Conv) and a VAE; the inclusion of mask in input (Mask), chosen from adding (w/) and not adding (w/o); the presence of cross-attention layers (Attn), chosen from adding (w/) and not adding (w/o); the type of UNet feature addition (UNet), selected from adding the full UNet feature (full), adding half of the UNet feature (half), and adding the feature like ControlNet (CN); and finally, the blending operation (Blend), chosen from not adding (w/o), direct pasting (paste), and blurred blending (blur). \textcolor{Red}{\textbf{Red}} is the best result.}
\scalebox{0.75}{
\setlength{\tabcolsep}{0.5mm}{
\begin{threeparttable}
{
\begin{tabular}{ccccc|ccc|ccc|c}
\toprule
\toprule
\multicolumn{5}{c|}{\textbf{Metrics}}  & \multicolumn{3}{c|}{\textbf{Image Quality}} & \multicolumn{3}{c|}{\textbf{Masked Region Preservation}} & \textbf{Text Align} \\
\midrule
\textbf{Enc} & \textbf{Mask} & \textbf{Attn} & \textbf{UNet} & \textbf{Blend} & \textbf{IR}$_{^{\times 10}}$$\uparrow$   & \textbf{HPS}$_{^{\times 10^2}}$$\uparrow$  & \textbf{AS}$\uparrow$   & \textbf{PSNR}$\uparrow$     & \textbf{MSE}$_{^{\times 10^2}}$$\downarrow$  & \textbf{LPIPS}$_{^{\times 10^3}}$$\downarrow$      & \textbf{CLIP Sim}$\uparrow$     \\ \midrule 
\cellcolor{LightRed} Conv & w/ & w/o & full & \cellcolor{LightRed} w/o & 11.05&26.23&6.55&14.89&37.23&64.54&26.76\\
VAE & \cellcolor{LightRed} w/o & w/o & full & \cellcolor{LightRed} w/o& 11.55 & 27.70 & 6.57 & 17.96 & 26.38 & 49.33 & 26.87\\
VAE & w/ & \cellcolor{LightRed} w/ & full & \cellcolor{LightRed} w/o& 11.25 & 27.62 & 6.56 & 18.69 & 19.44 & 34.28 & 26.63\\
\cellcolor{LightRed} Conv & w/ & \cellcolor{LightRed} w/ & \cellcolor{LightRed} CN & \cellcolor{LightRed} w/o&  9.58&26.85&6.47&12.15 & 80.91 & 150.89&26.88\\
VAE & w/ & \cellcolor{LightRed} w/ & \cellcolor{LightRed} CN & \cellcolor{LightRed} w/o& 10.53 & 27.42 & 6.59 & 18.28 & 24.36 & 41.63 & 26.89\\
VAE & w/ & w/o & \cellcolor{LightRed} CN & \cellcolor{LightRed} w/o& 11.42 & 27.69 & 6.58 & 18.49 & 24.09 & 36.33 & 26.86 \\
VAE & w/ & w/o & \cellcolor{LightRed} half & \cellcolor{LightRed} w/o& 11.47 & 27.70 & 6.57 & 19.01 & 23.77 & 33.57 & 26.87 \\
VAE & w/ & w/o & full & \cellcolor{LightRed} w/o & 11.59 & 27.71 & \textcolor{Red}{\textbf{6.59}} & 19.86 & 16.09 & 31.68 & \textcolor{Red}{\textbf{26.91}}\\
VAE & w/ & w/o & full & \cellcolor{LightRed} 
 paste &11.72&27.93&6.58&-&-&-&26.80\\
\midrule 
VAE & w/ & w/o & full & blur & \textcolor{Red}{\textbf{11.76}}&\textcolor{Red}{\textbf{27.94}}&6.58&\textcolor{Red}{\textbf{29.88}}&\textcolor{Red}{\textbf{1.53}}&\textcolor{Red}{\textbf{11.65}}&26.81 \\ 

     \bottomrule
     \bottomrule
\end{tabular}
}
\end{threeparttable}
}
}
\label{tab:ablation}
\end{table}

%% file: sections/conclusion.tex
\section{Discussion}
\label{sec:conclusion}

\paragraph{\textbf{Conclusion.}} This paper introduces a novel Inpainting-based Instruction-guided Image Editing paradigm (IIIE), which combines large language models (LLMs) and plug-and-play, all-in-one image inpainting models to enable autonomous, user-friendly, and interactive free-form instruction editing. Quantitative and qualitative results on PnPBench, our proposed benchmark, \Benchmark, and EditBench demonstrate the superior performance of \OurMethod~in terms of masked background preservation and image-text alignment in image editing and inpainting tasks.

\paragraph{\textbf{Limitations and Future Work.}} However, \OurMethod~has some limitations: (1) The quality and content generated by our model heavily depend on the selected base model. (2) Even with \OurMethod, poor generation results still occur when the mask has an irregular shape or when the provided text does not align well with the masked image. In future work, we aim to address these challenges.

\paragraph{\textbf{Negative Social Impact.}} Image inpainting models offer exciting opportunities for content creation but also present potential risks to individuals and society. Their reliance on internet-collected training data may amplify social biases, and there is a specific risk of generating misleading content by manipulating human images with offensive elements. To mitigate these concerns, responsible use and the establishment of ethical guidelines are essential, which will also be a focus in our future model releases.